\documentclass{article}

\usepackage[final]{neurips_2019}

\usepackage[utf8]{inputenc}
\usepackage[T1]{fontenc}
\usepackage{hyperref}
\usepackage{url}
\usepackage{booktabs}
\usepackage{amsfonts}
\usepackage{nicefrac}
\usepackage{microtype}
\usepackage{graphicx}
\usepackage{xcolor}
\usepackage{lipsum}
\usepackage{listings}
\usepackage{amsmath}

\title{
  Exploring Efficient Learning of Small BERT Networks with LoRA and DoRA
}

\author{
  Daniel Frees \\
  Department of Statistics \\
  Stanford University \\
  \texttt{dfrees@stanford.edu} \\
  \And 
  Aditri Bhagirath \\
  Department of Computer Science \\
  Stanford University \\
  \texttt{aditri@stanford.edu}\\
  \And
  Moritz Bolling \\
  Department of Statistics \\
  Stanford University \\
  \texttt{mbolling@stanford.edu} \\
}

\begin{document}

\maketitle

\begin{abstract}

While Large Language Models (LLMs) have revolutionized artificial intelligence, fine-tuning LLMs is extraordinarily computationally expensive, preventing smaller businesses and research teams with limited GPU resources from engaging with new research. \cite{hu2021lora}  and \cite{liu2024dora} introduce Low-Rank Adaptation (LoRA) and Weight-Decomposed Low-Rank Adaptation (DoRA) as highly efficient and performant solutions to the computational challenges of LLM fine-tuning, demonstrating huge speedups and memory usage savings for models such as GPT-$3$ and RoBERTa. We seek to expand upon the original LoRA and DoRA papers by benchmarking efficiency and performance of LoRA and DoRA when applied to a much smaller scale of language model: our case study here is the compact minBERT model. Our findings reveal that optimal custom configurations of LoRA and DoRA, coupled with Automatic Mixed Precision (AMP), significantly enhance training efficiency without compromising performance. Furthermore, while the parameterization of minBERT is significantly smaller than GPT-$3$, our results validate the observation that gradient updates to language models are inherently low-rank even in small model space, observing that rank $1$ decompositions yield negligible performance deficits. Furthermore, aided by our highly efficient minBERT implementation, we investigate numerous architectures, custom loss functions, and hyperparameters to ultimately train an optimal ensembled multitask minBERT model to simultaneously perform sentiment analysis, paraphrase detection, and similarity scoring.

\end{abstract}



\section{Introduction}
\vspace{-0.25cm}

In recent years, large language models have increased massively in terms of both performance and size. Current best estimations of large language model performance scaling laws, such as \cite{kaplan2020scaling} suggest that this trend will continue for the foreseeable future, with model performance scaling consistently as a power law relative to model size, dataset size, and the amount of compute used for training. As a consequence of the massive scale of modern large language models, the accessibility of language model fine-tuning has significantly decreased. 

The traditional approach for fine-tuning involves updating all model parameters, which can quickly become infeasible as the number of model parameters grows. However, this methodology came into question when \cite{li2018measuring} uncovered the promising finding that many models could be massively compressed into smaller parameter subspaces without compromising performance. \cite{aghajanyan2020intrinsic} extended this finding to demonstrate that fine-tuning updates to pre-trained large language models had low intrinsic rank. Subsequently, \cite{hu2021lora} and \cite{liu2024dora} proposed LoRA and DoRA, respectively, as solutions towards parameter efficient fine-tuning of pre-trained large language models, achieving a $10,000x$ reduction in trainable parameters and $3x$ reduction on GPU memory requirement when fine-tuning GPT-$3$ $175$B, without sacrificing performance. 

To the best of our knowledge, no prior research has benchmarked efficiency gains or implementation best practices when leveraging LoRA and DoRA for the fine-tuning of small BERT models. As an extension to the work of \cite{hu2021lora} and \cite{liu2024dora}, we seek to understand the intrinsic rank of fine-tuning updates, benchmark expected efficiency gains and performance reductions, and propose optimal selection of layers for low-rank adaptation of small BERT models. Our case study is our $\approx 226M$ (\autoref{sec:param_ct}) parameter minBERT model (built off of Stanford's CS224N minBERT: \cite{default_proj_handout}). To achieve this, we design custom implementations of LoRA and DoRA for minBERT and quantify efficiency gains and performance reduction for a multi-task minBERT model. Our model is designed to perform sentiment analysis, paraphrase detection, and semantic textual similarity scoring simultaneously. As a secondary goal, empowered by our efficient fine-tuning capabilities, we conduct extensive experiments towards deriving an optimal minBERT multi-task model.

\vspace{-0.25cm}

\section{Related Work}
\vspace{-0.25cm}

\subsection{Efficient Fine-Tuning}

\vspace{-0.25cm}
\cite{li2018measuring} uncovered the fact that many models can be re-parameterized in much lower dimensional space without losing performance. Building upon this, \cite{aghajanyan2020intrinsic} set out to investigate the intrinsic rank of fine-tuning updates for pre-trained large language models, finding that Li et al.'s results held true in the language modeling domain. Subsequently, \cite{hu2021lora} and \cite{liu2024dora} proposed parameter efficient fine-tuning techniques to take advantage of the low intrinsic rank in large language model fine-tuning updates. The novel contribution of \cite{hu2021lora}, Low-Rank Adaptation for Large Language Models (LoRA), has gained massive popularity in both academia and industry, enabling research teams with limited resources to explore fine-tuning use cases that might not otherwise be approachable. \cite{liu2024dora} was released earlier this year and suggests that LoRA tine-tuned model performance might be massively improved by a simple decomposition of weight updates into separate magnitude and direction components. 
 
LoRA's original benchmark results demonstrated a striking reduction in VRAM usag ——for instance, it reduced memory requirements from 1.2TB to 350GB during training on GPT-3 175B \cite{hu2021lora}. \cite{hu2021lora} also benchmarked LoRA with impressive results on GPT-2, RoBERTa, and DeBERTa and evaluated fine-tuned models across a variety of benchmarks, including GLUE, WikiSQL, SAMSum, and the E2E NLG Challenge. RoBERTa and DeBERTa were fine-tuned with LoRA and performance tested against the GLUE benchmark, containing several subtasks such as MNLI (Multi-Genre Natural Language Inference) [\cite{williams2018broad}], SST-2 (Stanford Sentiment Treebank) [\cite{socher2013recursive}], MRPC (Microsoft Research Paraphrase Corpus) [\cite{dolan2005automatically},] and QQP (Quora Question Pairs) [\cite{qqp2017}]. On these subtasks, LoRA achieved scores close to full fine-tuning, and even outperformed on some tasks. The LoRA adaptations of GPT-2 and GPT-3, tested on WikiSQL (natural language to SQL conversion) and E2E NLG (generating descriptions of restaurants from structured data), respectively, performed better than full fine-tuning while drastically reducing the number of parameters by up to 10,000 times \cite{hu2021lora}.  
\vspace{-0.25cm}

\subsection{Performance Optimizations}
\vspace{-0.25cm}

Several additional works guided our multi-task performance optimization experiments. The Adamax optimizer was proposed alongside the Adam optimizer by \cite{kingma2015adam}, where moment updates rely on the infinity norm, $L^p, p \rightarrow \infty$, rather than the $L^2$ norm. Given that we investigate fine-tuning of a small BERT model, our gradient updates involved high ($768$) dimensional embedding updates and potentially low intrinsic-rank attention layer updates. As such, the infinity norm may be a suitable measurement of gradient magnitude since our gradient updates are likely sparse. 

Recent works such as \cite{howard2018universal} have demonstrated the efficacy of iterative backbone unfreezing throughout fine-tuning of pre-trained large language models, whereby more and more of the large language model backbone is unfrozen as training progresses. Iterative unfreezing may yield superior generalization model performance and increase robustness of transfer learning to the phenomenon of catastrophic forgetting \cite{kirkpatrick2017overcoming}.

As recently confirmed by \cite{Mohammed2023EnsembleDeepLearning}, ensembling methods provide consistent performance gains and generally yield flatter, more generalizable optima. This fundamental principle applies across traditional regression/classification methods, tree-based methods, and deep learning, though in deep learning there is still some debate as to whether ensembling provides gains that can't be equivalently accomplished by larger deep neural network architectures. 

A portion of our experiments are dedicated to elucidating whether Adamax, iterative unfreezing, and ensembling are suitable approaches to increase the multi-task performance of small fine-tuned BERT models.

\vspace{-0.25cm}

\section{Approach}
\vspace{-0.25cm}

We implement all of the following model architectures and fine-tuning approaches ourselves using PyTorch (\cite{paszke2019pytorch}). 
\vspace{-0.25cm}

\subsection{Multi-task Baseline}
\label{subsec:multi_baseline}
\vspace{-0.25cm}

As a baseline, we implemented a minBERT model as described in Stanford's CS224N default project handout (\cite{default_proj_handout}), with $1D$ convolutions and $1$-layer fully-connected-network (FCN) classifier layers for each of our three tasks (sentiment, paraphrase detection, similarity scoring). Raw BERT embeddings from the minBERT backbone were fed as input into the SST classifier, while absolute BERT embedding differences between sentence pairs were computed and fed as input into the Quora and STS classifiers. Baseline training utilized a training routine where SST, Quora, and STS were each trained in turn, once per epoch, using Cross-Entropy (CE) loss for SST sentiment classification, Binary Cross-Entropy (BCE) loss for Quora paraphrase binary classification, and mean squared error (MSE) loss for similarity score prediction. 

\vspace{-0.25cm}

\subsection{Optimized Multi-task Model}
\label{sec:best_arch}
\vspace{-0.25cm}

\begin{figure}[h!]
  \centering
  \includegraphics[width=0.75\textwidth]{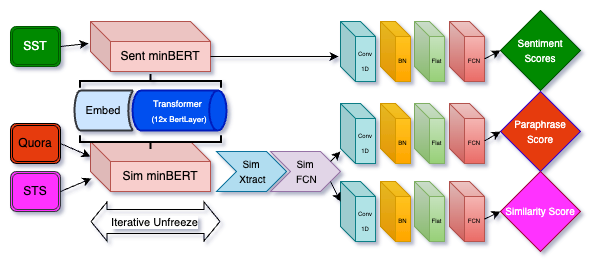}
  \caption{Architecture of our optimized minBERT model.}
  \label{fig:minbert_arch}
\end{figure}

Following the results of our performance optimization experiments (\autoref{subsec:perf_exp}), our optimized model architecture was constructed as depicted in \autoref{fig:minbert_arch}. Our BERT backbone consists of two identical but separate "Sent" and "Sim" pre-trained minBERT instances. A similarity feature extraction layer extracts features from sentence pair embeddings, concatenating the raw embeddings, $e_1, e_2$ with their elementwise product, $e_1 \odot e_2$, cosine similarity, $\frac{e_1 \cdot e_2}{\|e_1\| \|e_2\|}$, and absolute difference $\|e_1 - e_2\|$. All classifier layers consist of $1D$ convoltuions with kernel size $3$ and $4$ filters. BatchNorms are applied after convolutional and fully-connected layers, and we use ReLU non-linearity. Iterative unfreezing is applied to the backbone such that only a subset of parameters, $\theta + \omega$, are trainable at any given time. We denote classifier parameters as $\theta = \theta_{SST} + \theta_{Para} + \theta_{STS}$ and BERT backbone parameters as $\Omega$, with trainable backbone parameters $\omega \subseteq \Omega$.

\vspace{-0.25cm}

\subsection{Multi-task Learning Problem Formulation}
\vspace{-0.25cm}

Based on the results of our performance optimization experiments around loss functions (\autoref{sec:customloss}, we utilize Cross-Entropy loss to optimize for the SST sentiment task, Binary Cross-Entropy loss to optimize for the Quora paraphrase detection task, and a custom Pearson correlation loss to optimize for the STS similarity task. Thus we can consider our learning procedure to be an optimization problem where we seek to minimize the following composite loss

\[
L(\theta) = L_{\text{CE}}(f_{\theta_{\text{SST}} + \Omega}, y_{\text{SST}}) + L_{\text{BCE}}(f_{\theta_{\text{Para}} + \Omega}, y_{\text{Para}}) + L_{\text{Pearson}}(f_{\theta_{\text{STS}} + \Omega}, y_{\text{STS}})
\]
where \( f_{\theta_{\text{SST}} + \Omega}, f_{\theta_{\text{Para}} + \Omega}, f_{\theta_{\text{STS}} + \Omega} \) represent the model logits for SST, Quora paraphrase, and STS tasks respectively. We note also that our optimization problem has increasing degrees of freedom at each epoch depending how many of the backbone parameters $\omega \subseteq \Omega$ are unfrozen and trainable. 


\vspace{-0.25cm}

\subsection{Fine-Tuning Efficiency}
\vspace{-0.25cm}

\subsubsection{Baselines}
\vspace{-0.25cm}

Our baseline reference point for memory usage, time and accuracy consisted of the training baseline described in \autoref{subsec:multi_baseline} with all layers unfrozen, and without any automatic mixed-precision (AMP) or efficient fine-tuning (LoRA and DoRA) applied. We derived a lower bound for accuracy and an upper bound for efficiency by comparing against fine-tuning of only the classifier layers. 

\vspace{-0.25cm}

\subsubsection{AMP} 
\vspace{-0.25cm}

With automatic mixed precision (AMP) training, we computed forward passes and loss during training in half-precision format (\texttt{float16}), then cast back to full \texttt{float32} to backpropagate gradients, using a gradient scaler to minimize loss of gradient information due to \texttt{float16} underflow. This can substantially improve training throughput by reducing memory usage (\cite{nvidia_mixed_precision}). 
For an overview of the mathematical principles behind AMP, see \autoref{subsec:ampappendix}.
\vspace{-0.25cm}

\subsubsection{LoRA} 
\vspace{-0.25cm}

Various Python libraries, such as \texttt{peft} (\cite{peft}), provide implementations of LoRA and DoRA for common model architectures. However, we built a custom implementation tailored to the minBERT model, granting us precise control over the layers subjected to LoRA modifications. LoRA involves freezing the pre-trained weights of a linear layer ($W_0$) and integrating trainable low-rank matrices $B$ and $A$. This adaptation is represented by the equation:
\[
W' = W_0 + BA
\]
where $W_0 \in \mathbb{R}^{d \times k}$, $B \in \mathbb{R}^{d \times r}$, and $A \in \mathbb{R}^{r \times k}$ with $r \ll \min(d, k)$. This methodology reduces the number of parameters that need to be updated, thereby enhancing the efficiency of fine-tuning. While \cite{hu2021lora} introduced an additional parameter $\alpha$ that scales the weight update by $\frac{\alpha}{r}$, they acknowledged that using the Adam optimizer makes tuning $\alpha$ nearly equivalent to adjusting the learning rate. Consequently, we maintained $\alpha$ at a constant value equal to $r$ since we utilized Adam variants for efficiency experiments.

To implement LoRA, we froze all original parameters of each linear layer and initialize matrices $A$ and $B$ according to the layer's dimensions and the rank hyperparameter $r$, with $A$ receiving Gaussian initialization and $B$ being zero-initialized as described by \cite{hu2021lora}. During the forward pass, outputs from the original layer and the $BA$ product were computed and summed.
\vspace{-0.25cm}

\subsubsection{DoRA} 
\vspace{-0.25cm}

DoRA was proposed recently by Liu et al. (2024) \cite{liu2024dora} as a more performant version of LoRA, with weight updates decomposed into directional and magnitude components, offering more precise control over updates. The weights are updated as follows:
\[
W' = \frac{m}{\|W_0 + BA\|_c} (W_0 + BA)
\]
where $\Delta V$ represents the incremental directional update, $m = \|W_0\|_c$, and $\|\cdot\|_c$ denotes the vector norm of a matrix across its columns. Note that $m$ is a  trainable magnitude parameter ($m$) initialized to the column norms of the original weights. While \texttt{peft} recently released initial implementations of DoRA, we implemented DoRA from scratch based loosely on the work of \cite{dora2023github}.
\vspace{-0.25cm}

\subsubsection{LoRA and DoRA Modes}
\vspace{-0.25cm}

Our custom implementations of LoRA and DoRA enabled four distinct matrix decomposition modes:

\begin{enumerate}
    \item \textbf{attn-only}: This mode aligns with the recommendations from \cite{hu2021lora}, applying LoRA only to the attention mechanism's query, key, and value matrices while freezing all other parameters except classifier layers. Especially powerful for transformer architectures.
    \item \textbf{attn}: Similar to the \textit{attn-only} mode, this setting replaces the attention matrices (query, key, and value) but does not freeze the remaining layers.
    \item \textbf{all-lin-only}: In addition to query, key, and value matrices, this mode extends LoRA to all linear layers within the minBERT model. All other layers except classifiers are frozen. 
    \item \textbf{all-lin}: Operates similarly to \textit{all-lin-only} by applying LoRA across all linear layers, but does not freeze the other layers, permitting full model dynamics during training.
\end{enumerate}

These modes offer varying degrees of parameter adaptability and training dynamics, enabling the tailored application of LoRA/DoRA depending on specific model requirements and targeted results. In terms of the number of parameters being updated, \textit{attn-only} modifies the fewest, while \textit{attn} adjusts the most. As a reference point, the entire minBERT model has 110M parameters. 21M of these are contained in linear layers within the attention blocks, a further 64M are contained in linear layers outside of the attention blocks.



\vspace{-0.25cm}

\subsection{Performance Optimization} 
\vspace{-0.25cm}

Our efficient fine-tuning techniques enabled us to perform numerous experiments towards deriving the most performant multi-task minBERT model possible. We experimented with Linear, NonLinear, and Convolutional classifiers, cosine annealing and multiplicative learning rate scheduling, exponential moving average (EMA) weight updates, custom loss functions (Pearson loss, hybrid MSE/BCE loss, ordinal CE loss, SimCSE contrastive loss and others), various data cleaning pipelines, various similarity task feature extraction layers, iterative BERT backbone unfreezing, various optimizers (SGD, AdamW, RAdam, Adamax, SparseAdam) and task-specific optimizers, weight decay strength, initial learning rate selection, sub-task learning rate multipliers, multiple interleaving training schedules, and finally model ensembling and optimal ensemble mixtures. 
\vspace{-0.25cm}

\section{Experiments}
\vspace{-0.25cm}

All experiments were conducted on NVIDIA A100 Tensor Core GPU with 40GB of VRAM. We conducted 3 main experiments to evaluate the impact of our efficiency improvements.
\vspace{-0.25cm}

\subsection{Data}
\vspace{-0.25cm}

We experiment with Stanford Sentiment Treebank (SST), Quora Question Pairs (QQP), and Semantic Textual Similarity Benchmark (STS) datasets for our three model sub-tasks. Sub-tasks and corresponding dataset details are included in \autoref{tab:datatable}.

\begin{table}[h]
    \small
    \begin{tabular}{@{}p{2cm}p{6cm}p{6cm}@{}}
        \toprule
        \textbf{Task} & \textbf{Input} & \textbf{Output} \\
        \midrule
        \textbf{Sentiment Analysis (SST)} & \textbf{$11,855$
sentences} from movie reviews.& A \textbf{sentiment score} ranging from 0 (Negative) to 4 (Positive)\\
        \addlinespace
        \textbf{Paraphrase Detection (QQP)} & \textbf{$404,298$ question pairs} with binary labels indicating whether the questions are paraphrases. & A \textbf{binary label (Yes or No)}, indicating whether the question pairs are paraphrases. \\
        \addlinespace
        \textbf{Semantic Textual Similarity (STS)} & \textbf{$8,628$ sentence pairs} with similarity scores. & A \textbf{semantic similarity score} on a scale from $0$ (unrelated) to $5$ (equivalent meaning) \\
        \addlinespace
        \bottomrule
    \end{tabular}
    \caption[Tasks and Data Overview]{Overview of Data and Tasks}
    \label{tab:datatable}
\end{table}

We perform some basic data preprocessing before passing inputs into our multi-task BERT model (see \autoref{sec:data_preprocessing} for details). 
\vspace{-0.25cm}

\subsection{Evaluation methods}
\vspace{-0.25cm}

We evaluate our models using multi-class acurracy for sentiment analysis (SST), binary accuracy for paraphrase detection (Quora) and Pearson correlation for semantic textual similarity (STS). For a total multi-task performance score (and for the purposes of checkpointing our best model weights during training), we adopt the same formula as the CS224N project competition leaderboard, normalizing the Pearson correlation to a $[0,1]$ range and taking a simple average across the three tasks.

Additionally, to assess the efficiency of our models, we record the average time and average peak memory usage per training epoch for each task. Due to significant variability in computational demands across tasks——with the Quora paraphrase detection task requiring approximately $15$ times more resources than the SST and STS tasks——these measures are averaged on a per-task basis. 
\vspace{-0.25cm}

\subsection{Performance Experiments and Results}
\label{subsec:perf_exp}
\vspace{-0.25cm}

After initial experiments on optimizers (\autoref{sec:adamax}), learning rate, weight decay, architecture (\autoref{sec:chron_exp}), iterative unfreezing (\autoref{sec:unfreezing}), EMA weight updates (\autoref{sec:ema}), custom loss functions (\autoref{sec:customloss}), ensembling (\autoref{sec:ensembling}) and optimal ensemble mixtures of models with varied sub-task bias (\autoref{sec:final_ensembles}, our optimal model architecture was implemented as reported in \autoref{sec:best_arch}, and our best final model consists of $23$ different models ensembled together, across our "G6 Duality of Man" (\autoref{sec:duality_of_man}), "G6 Yin Yang" (\autoref{sec:yinyang}), and "G6 Interleaver" (\autoref{sec:interleaver_models} model architectures/ learning routines. We dub this best ensemble "G6 Lisan Al Gaib", and report its results on the test dataset in \autoref{tab:results}. We also report our "G6 Oligarch v1" ensemble results.

\begin{table}[h]
\centering
\begin{tabular}{lcccc}\toprule
Model & Sentiment & Paraphrase & Similarity & Overall Score \\ \midrule
\textbf{G6 Oligarch v1}  & \textbf{0.551} & 0.881 & 0.853 & 0.786 \\
\textbf{G6 Lisan al Gaib}  & 0.543 & \textbf{0.889} & \textbf{0.861} & \textbf{0.788} \\ \bottomrule
\end{tabular}
\caption{Best Ensemble Model Test Results}
\label{tab:results}
\end{table}
\vspace{-0.25cm}

\subsection{Efficiency Experimental Details and Results}
\vspace{-0.25cm}

\subsubsection{Large Grid Search}
\vspace{-0.25cm}

To evaluate the impact of LoRA and DoRA on our multitask classifier, we conducted a comprehensive grid search over the parameters summarized in Table \autoref{subsec:grid_search_details}.

\begin{figure}[ht]
    \centering
    \includegraphics[scale=0.5]{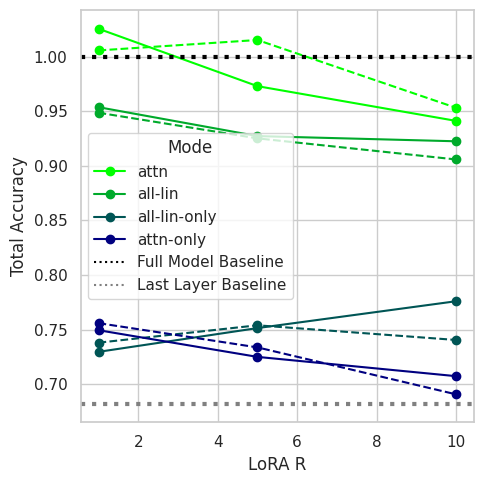}
    \includegraphics[scale=0.5]{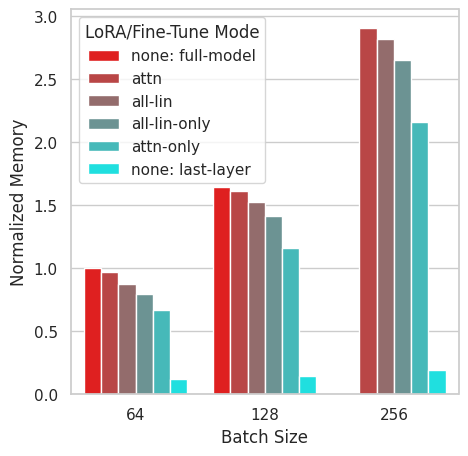}
    \caption{Left: LoRA and DoRA Accuracy per Mode across Rank. Dashed lines represent DoRA, solid lines LoRA. Normalized to Full Model baseline. Right: Normalized Peak Memory Consumption compared to Batch Size.}
    \label{fig:gridsearch_acc_v_R}
\end{figure}

The results of this grid search are presented in \autoref{fig:gridsearch_acc_v_R}, where values are averaged over all parameters and for each decomposition rank $r$. We observe a decrease in total accuracy with fewer fine-tunable parameters, as expected. Moreover, examining the results by batch size, memory consumption exhibits the anticipated trend: greater numbers of parameters correspond to higher memory usage, while fewer parameters result in reduced memory consumption. However, we also make two unexpected observations: 1) accuracy does not improve with increasing rank; and 2) DoRA does not outperform LoRA. These findings suggest that the intrinsic dimensionality of the weight updates for our small minBERT model are sufficiently low such that most information can be captured by a rank-one approximation. Higher ranks or distinctions between directionality and magnitude may capture only minimal additional nuances, overshadowed by the variability introduced. This hypothesis was substantiated through a comprehensive series of experiments comparing LoRA and DoRA across various ranks, which demonstrated that while memory usage and computational time escalate with rank, accuracy does not. Detailed results and analysis are provided in \autoref{subsec:lora_dora_many_r}.

\vspace{-0.25cm}

\subsubsection{Ablation Study}
\vspace{-0.25cm}

To gauge the memory, time, and performance impact of various combinations of LoRA, DoRA, and AMP, we ran an ablation experiment, collecting metrics across all configurations. The hyperparameters for the experiment setup and the detailed results can be found in \autoref{subsec:ablation_details}. In summary, the key findings are that AMP led to a reduction of memory usage by 26\%, combining it with LoRA \textit{attn-only} brought this to a 50\% reduction, while using LoRA \textit{attn} brought it to 28\%. We noticed that in one case, using AMP caused the accuracy to drop significantly. This is likely due to our baseline efficiency experiment model being unstable (this model did not yet use Adamax and multiple minBERT backbones). That said, even though we apply a gradient scaler to avoid underflow, AMP can increase model instability in that we lose some precision in our gradient update space. In general AMP left accuracy unaffected. AMP yielded significant average time speedups of 60\%, whereas LoRA and DoRA caused negligible time improvements. These results are as expected as LoRA and DoRA's main advantage is the memory savings they bring.
\vspace{-0.25cm}

\subsubsection{Applying Improvements to Final Model}
\vspace{-0.25cm}

\begin{figure}[ht]
    \centering
    \includegraphics[scale=0.3]{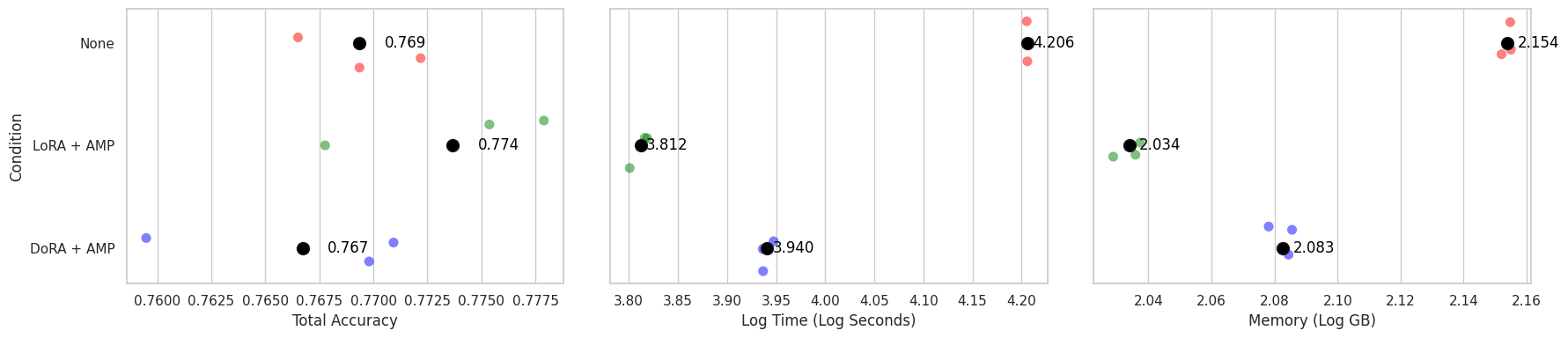}
    \caption{Comparison of Total Accuracy, Average Log Time and Average Log Memory across all tasks. The black dots represent the average over all 3 runs.}
    \label{fig:final_model_acc}
\end{figure}

Finally, we evaluated the performance of our best (non-ensemble) model architecture (\autoref{sec:best_arch}) with and without efficiency optimizations, employing AMP with either LoRA or DoRA in the \textit{all-lin} mode, across three seeds and seven epochs of training. The configurations are detailed in \autoref{tab:final_model_benchmark_hyperparams} and the accuracy results in \autoref{fig:final_model_acc}. The LoRA + AMP combination slightly outperformed the baseline, while DoRA + AMP marginally underperformed, with variances within a 1\% range suggesting no statistical significance. Despite these minor performance differences, LoRA + AMP was more resource-efficient, showing a 1\% improvement in performance, 33\% faster execution, and 12\% less memory usage than the baseline. Conversely, DoRA + AMP, despite a slight performance dip, was 24\% faster and used 7\% less memory. The memory savings were somewhat muted due to the iterative unfreezing approach and averaging our measurements over all epochs.

\vspace{-0.25cm}

\section{Analysis }
\label{sec:analysis}
\vspace{-0.25cm}

Our findings suggest that with optimal hyperparameters, the integration of LoRA and DoRA with AMP can significantly enhance memory and time efficiency with minimal impact on model accuracy. We observed memory reductions ranging from 7\% to 50\% and time efficiency improvements up to 60\%. Notably, the effect on accuracy does not correlate with LoRA rank but is highly dependent on its operational mode, with observed variations from a decrease of 24\% to an increase of 3\%. Low-rank adaptations, due to their reduced memory demands and relatively modest performance degradation, are particularly apt for scenarios requiring frequent model updates, such as real-time language processing or environments constrained by hardware capabilities. Furthermore, our study underscores the necessity for a nuanced understanding and thorough evaluation of potential failure cases induced by these adaptations, particularly in high-stakes scenarios. Before deployment, it is crucial to assess whether these adaptations introduce undesirable biases or failure patterns. Additionally, we found that STS tasks suffer an average performance decrease of 4\% more compared to sentiment classification tasks, suggesting task-specific changes in the intrinsic dimension of the weight updates.

For sentiment analysis, outputs from LoRA and DoRA closely matched those of our best model. In scenarios where DoRA predictions were incorrect, they deviated from our optimal model's predictions by only one point in 94\% of the cases. This variance is considered acceptable given the subjective nature of sentiment analysis. Examples where DoRA's sentiment predictions significantly diverged from our best model are presented in Table \autoref{tab:sentiment_discrepancies}, typically involving sentences with mixed emotional content, which seem to have posed challenges for DoRA in capturing nuanced sentiment.

\begin{table}[h!]
    \centering
    \caption{Example Discrepancies in Sentiment Score Prediction: DoRA vs. Our Best Model}
    \label{tab:sentiment_discrepancies}
    \begin{tabular}{|p{5.5cm}|c|c|c|}
    \hline
    \textbf{Sentence} & \textbf{True Sentiment} & \textbf{DoRA Score} & \textbf{Best Model Score} \\
    \hline
    ``It's one pussy-ass world when even killer-thrillers have no edge.'' & 0 & 3 & 1 \\
    \hline
    ``It's like watching a nightmare made flesh.'' & 0 & 3 & 1 \\
    \hline
    ``Harland Williams is so funny in drag he should have his own TV series.'' & 2 & 3 & 1 \\
    \hline
    \end{tabular}
\end{table}

In terms of paraphrase detection, our model attained a peak accuracy of 0.888, with the DoRA version achieving a comparable 0.858. The performance of DoRA on the STS task was also close to that on sentiment classification, mostly differing by minor points.

Regarding our secondary goal of deriving an optimal minBERT-based model for our multi-task reasoning benchmarks, we found that using \texttt{Adamax} as an optimizer (\autoref{sec:adamax}), strong multiplicative learning rate decay ($0.5^{\text{epoch}} \times LR$), separate Sentiment and Similarity minBERT backbones (\autoref{sec:duality_of_man}), shared Quora and STS non-linear feature extraction layer, $1D$ convolutional classifiers, Pearson correlation loss (\autoref{sec:customloss}), and ensembling of as many models as possible (\autoref{subsec:ensemble_results}) were critical to maximizing performance. 


\vspace{-0.25cm}

\section{Conclusion }
\vspace{-0.25cm}


We extend the work of \cite{hu2021lora} and \cite{liu2024dora} by customizing LoRA and DoRA for minBERT, and benchmarking efficiency and accuracy effects of low-rank adapation on small BERT models. Our results confirm that even for small BERT models the intrinsic dimension of the weight updates, $\Delta W$, is $1$. Increasing the rank beyond this point does not yield significant performance gains. The fact that \texttt{Adamax} proved to be our best optimizer may be further evidence that the intrinsic rank of our fine-tuning updates was very low, given that \texttt{Adamax} is well-suited to sparse gradient updates.

We find that for our small BERT model, LoRA provides sufficient granularity for performant fine-tuning, with LoRA fine-tuning exceeding vanilla training performance. Our DoRA fine-tuning approach achieves slightly lower performance, suggesting that we do not benefit from separately decomposing magnitude and direction for our weight updates. Moreover, while the efficiency gains demonstrated here on minBERT are more modest compared to those observed in larger models such as GPT-3, our findings endorse the adoption of AMP combined with LoRA, applied to all linear transformer layers, even within smaller language model frameworks. These efficiency optimizations facilitate broader architectural explorations and the utilization of computationally expensive techniques like ensembling, enhancing the model's performance and feasibility for extensive deployment. While we make qualitative observations about model errors when using parameter efficient fine-tuning versus full fine-tuning (see \autoref{sec:analysis}), further research is necessary to elucidate the exact relationship between low-rank weight updates to small BERT models and the specific types of errors made by these models on downstream tasks.


\clearpage

\section{Ethics Statement}

This work analyzes minBERT, a compressed version of BERT. As a result of its decreased parameterization, our minBERT model may not capture the full richness and complexity of the data it was trained on. For instance, minBERT may not capture the appropriate use of terms based on social context, varied dialects or regional differences, and may generate responses that are too reductive based on general patterns observed in the data. To mitigate this, we propose conducting comprehensive bias audits [\cite{dwt_2024}] to identify any problematic patterns in task predictions, and involving community stakeholders in an iterative model refinement process prior to deployment in any high-risk use cases, such as chat applications, social media, loan approval, etc. Incorporating feedback from diverse user groups and integrating domain-expert rules can help address specific biases and ensure more equitable model performance.

Using compression techniques such as LoRA and DoRA may also result in a loss of interpretability of our model. By introducing low-rank matrices which interact with existing high-dimensional parameter spaces, we further obfuscate the model's decision-making techniques. While deep-learning models are already challenging to interpret, techniques such as attention heat-mapping can be used to gain qualitative insight into model predictions. No studies to date have specificially investigated the impact of low-rank fine-tuning on interpretive visualizations, so it is entirely possible that we lose attention granularity. To combat this loss in interpretabality, future research should explore designing visual tools to understand how adjusting intermediate matrices influences model outputs, using high-rank update counterfactuals. This may offer a clearer picture of the impact of these compression techniques on model decision-making. Furthermore, establishing new interpretability benchmarks specific to models using LoRA and DoRA, combined with controlled A/B testing in deployment, can provide insights into the practical impacts of these techniques and safeguard against unforeseen adverse outcomes. Interpretability is absolutely critical to any applications where our model might be deployed as a decision-maker, or incorporated into a decision making pipeline (e.g. loan applications, mortgage pricing). 

\bibliographystyle{acl_natbib}
\bibliography{efficientsmalldora}

\clearpage

\appendix

\section{Adamax Optimizer}
\label{sec:adamax}

The Adamax optimizer was proposed alongside the Adam optimizer by \cite{kingma2015adam}, where moment updates rely on the infinity norm, $L^p, p \rightarrow \infty$, rather than the $L^2$ norm. For our use-case whereby we seek to propagate gradient updates to embeddings when training in \texttt{iterative} and \texttt{full-model} fine-tuning modes, the infinity norm may be a more suitable measurement of gradient magnitude, since our embeddings are high dimensional (size $768$) and gradient updates are likely sparse. Empirically, we found a significant performance increase when using Adamax as opposed to Adam and SGD (see \autoref{fig:adamax_better}). Adam and RAdam were also inferior to Adamax for our training process.

\begin{figure}[!ht]
    \centering
    \includegraphics[width=0.5\linewidth]{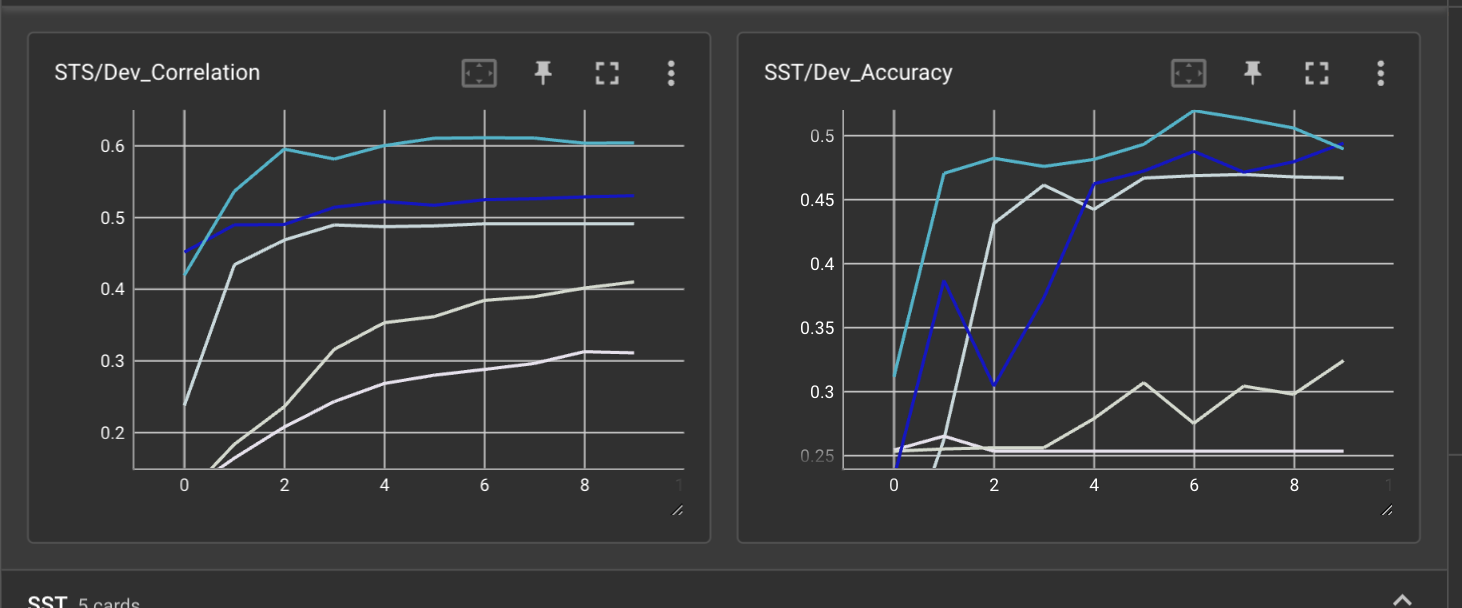}
    \caption{Adamax yields smoother performance increase and better convergence (light blue) as opposed to Adam (grey) and SGD (dark blue).}
    \label{fig:adamax_better}
\end{figure}

\clearpage

\section{Iterative BERT Backbone Unfreezing}
\label{sec:unfreezing}

Recent works such as \cite{howard2018universal} have demonstrated the efficacy of iterative backbone unfreezing throughout fine-tuning of pre-trained large language models, whereby more and more of the large language model backbone is unfrozen as training progresses. Iterative unfreezing may yield superior generalization model performance and increase robustness of transfer learning to the phenomenon of catastrophic forgetting \cite{kirkpatrick2017overcoming}. Motivated by the promising results of progressive unfreezing in the literature, we implement iterative unfreezing (see \autoref{subsec:unfreeze_code}) of the BERT backbones in our multit-task BERT model. Our early experiments comparing training convergence between full unfreezing and iterative unfreezing clearly demonstrated better convergence using the iterative strategy, and as such we sought to derive an optimal unfreezing schedule. Ultimately, our optimal schedule required fairly rapid unfreezing. We unfreeze $3$ additional BERT backbone attention layers (there are $12$ in total) each epoch until the entire backbone is unfrozen. Note also that when we unfreeze the first backbone layers, we make sure to unfreeze all post-attention BERT layers. Similarly, when all BERT backbone layers are unfrozen, we unfreeze not only through the first attention layer, but all the way back to the original Embedding layer. For a closer look at the iterative unfreezing controller, see \autoref{subsec:unfreeze_controller}.

\subsection{Flexible Unfreezing MultitaskBERT Class Method}
\label{subsec:unfreeze_code}

\begin{lstlisting}[language=Python, caption=Manage Freezing Function]
def manage_freezing(self, structure: str):
    """ 
    Manages freezing of BERT Attention layers

    Adjusts the trainable status of layers in a model based on a specified
    structure command. This function can freeze all layers, unfreeze all layers,
    or unfreeze only the top N layers of the model.

    Args:
        structure (str): A command string that dictates how layers should be
                         frozen or unfrozen. It can be 'freezeall', 'unfreezeall',
                         or 'unfreezetopN' where N is an integer indicating the
                         number of top layers to unfreeze.

    Raises:
        ValueError: If the structure parameter does not follow the expected
                    format or specifies an invalid option.
    """
    for bert in (self.bert_sentiment, self.bert_sim):
        children = list(bert.children())
        # The ModuleList containing the 12 self-attention layers is at index 5
        attn_idx = 5
        attention_layers = children[attn_idx]
        total_attn_layers = len(attention_layers)
        
        if structure == "freezeall":
            # Freeze all layers
            for param in bert.parameters():
                param.requires_grad = False

        elif structure == "unfreezeall":
            # Unfreeze all layers
            for param in bert.parameters():
                param.requires_grad = True

        elif structure.startswith("unfreezetop"):
            # Attempt to extract the number of attn. layers to 
            # unfreeze from the structure string
            try:
                n_layers = int(structure[len("unfreezetop") :])
            except ValueError:
                raise ValueError(
                    (
                        "Invalid layer specification. Ensure it "
                        "follows 'unfreezetopN' "
                        "format where N is a number."
                    )
                )

            # Freeze all layers first
            for param in bert.parameters():
                param.requires_grad = False
                
            # Unfreeze children after the attention layers 
            for i in range(attn_idx+1, len(children)):
                for param in children[i].parameters():
                    param.requires_grad = True
            # Unfreeze the last n attention_layers
            for i in range(total_attn_layers - n_layers, total_attn_layers):
                for param in attention_layers[i].parameters():
                    param.requires_grad = True
        else:
            raise ValueError(
                (
                    "Invalid structure parameter. Use 'freezeall',"
                    "'unfreezeall', or "
                    "'unfreezetopN' where N is a number."
                )
            )
    return None
\end{lstlisting}

\subsection{Iterative Unfreezing Controller for Training}
\label{subsec:unfreeze_controller}

\begin{lstlisting}
for epoch in range(args.epochs):
    if args.fine_tune_mode == 'iterative':
        checkpoint_slice = args.epochs // 5   # no longer using this, 
                            #but can be used for 'even-slice' unfreezing
        if epoch < 1:
            model.manage_freezing('freezeall')
            print(f"\n====== All BERT Layers Frozen ======\n")
        elif epoch < 2:
            model.manage_freezing('unfreezetop3')
            print(f"\n====== Top 3 BERT Layers Unfrozen ======\n")
        elif epoch < 3:
            model.manage_freezing('unfreezetop6')
            print(f"\n====== Top 6 BERT Layers Unfrozen ======\n")
        elif epoch < 4:
            model.manage_freezing('unfreezetop9')
            print(f"\n====== Top 9 BERT Layers Unfrozen ======\n")
        else:
            model.manage_freezing('unfreezeall')
            print(f"\n====== All BERT Layers Unfrozen ======\n")   
            # 12 total attn layers and early layers including Embeddings
\end{lstlisting}
\clearpage

\section{AMP}
\label{subsec:ampappendix}

There are a few key strategies that AMP employs to reduce computational load and memory utilization, primarily by using FP16 instead of FP32 wherever feasible, without sacrificing numerical stability and accuracy too significantly. This leads to performance improvements on hardware designed to effectively exploit lower precision arithmetic (such as our NVIDIA A100s).
\subsection*{Loss Scaling (LS)}

To prevent numerical underflow in lower precision (FP16), gradients are scaled up by a factor \( S \) during computation:
    \[
    L' = S \cdot L
    \]
After backpropagation, gradients are scaled down to update weights. This ensures that no significant digits are lost.
    \[
    \frac{\partial L}{\partial w} = \frac{1}{S} \cdot \frac{\partial L'}{\partial w}
    \]

\subsection*{Precision Casting}

Critical operations such as gradient accumulations and updates are performed in higher FP32 precision. To account for mixed precision operations, we cast appropriately to prevent information loss.

\subsection*{Master Weight Copy}

A high precision (FP32) copy of the weights is maintained. This ensures accurate update steps:
    \[
    w_{\text{new}} = w - \eta \cdot \text{cast}(\frac{\partial L}{\partial w}, \text{FP32})
    \]

\clearpage

\section{Ensembling}
\label{sec:ensembling}

Towards achieving optimal multi-task model performance, we ensemble various mixtures of high-performance models together. As recently confirmed by Mohammed and Kora (2023) \cite{Mohammed2023EnsembleDeepLearning}, ensembling methods provide consistent performance gains and generally yield flatter, more generalizable optima. This fundamental principle applies across traditional regression/classification methods, tree-based methods, and deep learning, though in deep learning there is still some debate as to whether ensembling provides gains that can't be equivalently accomplished by larger deep neural network architectures. In general, the two overarching ensemble approaches are to concatenate the predictions of many weak models, such that each learns from each others mistakes (e.g. XGBoost \cite{Chen2016XGBoost}) or to average the predictions of many strong models (e.g. RandomForest \cite{Breiman2001RandomForest}). 

In the present work, we apply the latter strategy to average the strengths and weaknesses of our various strong learners. We find that ensembling consistently increases performance across all three sub-tasks, and propose that despite the debate in deep learning literature as to whether ensembling provides gains inachievable with better architecture, the ease of ensembling versus spending days searching for more optimal architecture is well worth it when computational resources are sufficient to perform the ensembling. Given our general approach in this paper towards deriving more efficient, yet performant multi-task minBERT models, our ensembling approach can be regarded as a case study in the performance benefits that efficient fine-tuning can provide, agnostic of resource shortages. 

Our ensembling approach is as follows. We initialize an EnsembledModel class with a list of pre-loaded, trained MultitaskBERT models (or MultitaskBERTDualityOfMan for the older Duality of Man architecture). Then, we aggregate sentiment, paraphrase and similarity predictions by stacking predictions from all EnsembledModel sub-models into a tensor, and taking the mean of these predictions. In this way, we hope to approximate a better optimum point by averaging the predictions of a number of sub-models converged to less optimal local minima. 

Towards creating an ensemble which comprises of sub-models that excel at different sub-tasks, we implement separate optimizers with varying weight-decay and learning rate for each sub-task. We found that within a reasonable learning rate range ($\approx 1e-4$), variations in learning rate of up to $10x$ primarily had the effect of biasing model performance towards the task with larger learning rate. As such, we apply various learning rate scales to different sub-task across our ensemble sub-models (for examples, see the training invocations in \autoref{sec:duality_of_man}
and \autoref{sec:yinyang}). \\

Our EnsembledModel class implementation is as follows:

\begin{lstlisting}[language=Python, caption=Ensembled Model and Model Loading in PyTorch]
import torch

class EnsembledModel(nn.Module):
    def __init__(self, models):
        super(EnsembledModel, self).__init__()
        self.models = nn.ModuleList(models)

    def forward(self, input_ids, attention_mask):
        raise NotImplementedError("EnsembledModel does not support the forward "
            "method directly. Use predict_sentiment, predict_paraphrase, "
            "or predict_similarity methods.")

    def predict_sentiment(self, input_ids, attention_mask):
        with torch.no_grad():
            logits = torch.stack([model.predict_sentiment(input_ids, 
                                                    attention_mask) \
                                for model in self.models], dim=0)
            return torch.mean(logits, dim=0).squeeze()

    def predict_paraphrase(self, input_ids_1, attention_mask_1, 
                        input_ids_2, attention_mask_2):
        with torch.no_grad():
            logits = torch.stack([model.predict_paraphrase(input_ids_1, 
                                        attention_mask_1, 
                                        input_ids_2, 
                                        attention_mask_2) \
                                        for model in self.models], dim=0)
            return torch.mean(logits, dim=0).squeeze()

    def predict_similarity(self, input_ids_1, attention_mask_1, 
                        input_ids_2, attention_mask_2):
        with torch.no_grad():
            logits = torch.stack([model.predict_similarity(input_ids_1, 
                                            attention_mask_1, 
                                            input_ids_2, attention_mask_2) \
                                            for model in self.models], dim=0)
            return torch.mean(logits, dim=0).squeeze()
\end{lstlisting}

To load models into an EnsembledModel (with the desired filepaths in args.filepaths), a minimal example is as follows. For example, we call our overall ensembling script with \texttt{\$python ensemble.py model1.pt model2.pt}. 

\begin{lstlisting}
from multitask_bert import MultitaskBERT, MultitaskBERTDualityOfMan

def load_models(filepaths, device):
    models = []
    for filepath in filepaths:
        print(f"====== Loading Model: {filepath} ======")
        saved = torch.load(filepath, map_location=device)
        config = saved['model_config']
        if "duality" in filepath:
            model = MultitaskBERTDualityOfMan(config)
        else:
            model = MultitaskBERT(config)
        model.load_state_dict(saved['model'])
        model = model.to(device)
        model.eval()
        models.append(model)
    return models

def main(args):
    device = torch.device('cpu')
    if torch.cuda.is_available():
        device = torch.device('cuda')
    elif torch.backends.mps.is_available():
        device = torch.device('mps')

    model_list = load_models(args.filepaths, device)
    ensemble_model = EnsembledModel(model_list).to(device)

    test_multitask(args, model=ensemble_model) 
\end{lstlisting}

\subsection{Ensemble Results}
\label{subsec:ensemble_results}

We found ensembling to consistently boost performance (see \autoref{sec:duality_of_man}, \autoref{sec:yinyang}, \autoref{sec:runitback_model}, \autoref{sec:interleaver_models}), and in general more models yielded better performance (\autoref{sec:final_ensembles}). Notably adding more "votes" to better models had a very minor postive impact on performance (see Oligarchy v1 in \autoref{sec:final_ensembles}), but generally the effect of adding more models was stronger. Our best model, "G6 Lisan al Gaib" outperforms the "G6 The Sage" model, with the difference between them being that the "Lisan al Gaib" variant includes more relatively low-performance models. Even in the case of weaker sub-models, we note very good ensembled performance (e.g. \autoref{sec:interleaver_models}). 

\clearpage

\section{Exponential Moving Average (EMA) Model}
\label{sec:ema}

\subsection{Motivation}

We implement a multitask-compatible version of the exponential moving average (EMA) model, a form of Polyak averaging \cite{PolyakAveraging}. This decision was motivated by the success of stochastic weight averaging (SWA) introduced by  Izmailov et al. (2019) \cite{Izmailov2019AveragingWeights}, who propose SWA as a solution to achieve more optimal convergence using exponential moving averages and cyclical (or constant) learning rates as compared to traditional SGD with learning rate decay. The general proposal of SWA is that SGD solutions are often inoptimal, but an average of solution points yields a more generalizable optimum point that yields superior performance on unseen data. EMA is similar to SWA, except that a running average of weights is maintained and updated every epoch, rather than updating weights to be an average of weights from the 'converged' weights at the end of various cycles. 

\subsection{Adopted Method and Extensions}

We use a custom cyclical learning rate for our EMA experiments, motivated by the fact that Izmailov et al. (2019) find cyclically decreasing learning rates to better approximate solutions each cycle, despite the fact that the loss space is overall better explored by constant learning rate SGD due to its larger updates and greater movement. Given our compute limitations, we opted for cyclical learning rate scheduling, towards producing an average of fewer better weights versus the average of many weaker weights. 

 We implement a custom cyclical learning rate scheduler which decreases learning rate by \texttt{8x} each cycle, with cycles each running for $2$ epochs. We further match these $2$ epoch cycles up with our iterative unfreezing strategy, whereby each successive deepening of the unfreezing corresponds perfectly to a learning rate cycle. Because of the demonstrably superior performance of \texttt{Adamax} as an optimizer (see \autoref{sec:adamax}), we use \texttt{Adamax} as opposed to \texttt{SGD}, which is typically utilized for SWA methods. Thirdly, as an extension to the official PyTorch implementation of EMA, we design a custom implementation of the EMA model wrapper which performs weight-averaging each epoch as usual, but at evaluation-time adopts the multiple task output methods of our superclass \texttt{MultitaskBERT} model. We update BatchNorms at the end of training in line with the original work, using the official PyTorch implementation from \texttt{torch.optim.swa\_utils}.

\begin{lstlisting}[language=Python, caption=EMA Class]
import copy

class EMA:
    def __init__(self, model, decay):
        self.model = model
        self.decay = decay
        self.shadow = copy.deepcopy(model.state_dict())
        self.ema_model = copy.deepcopy(model)

    def update(self):
        for name, param in self.model.state_dict().items():
            if name in self.shadow:
                self.shadow[name] = self.decay * self.shadow[name] + \
                    (1.0 - self.decay) * param

    def apply_shadow(self):
        self.ema_model.load_state_dict(self.shadow)
        return self.ema_model
\end{lstlisting}

\subsection{EMA Results}

\begin{figure}[!ht]
    \centering
    \includegraphics[width=0.5\linewidth]{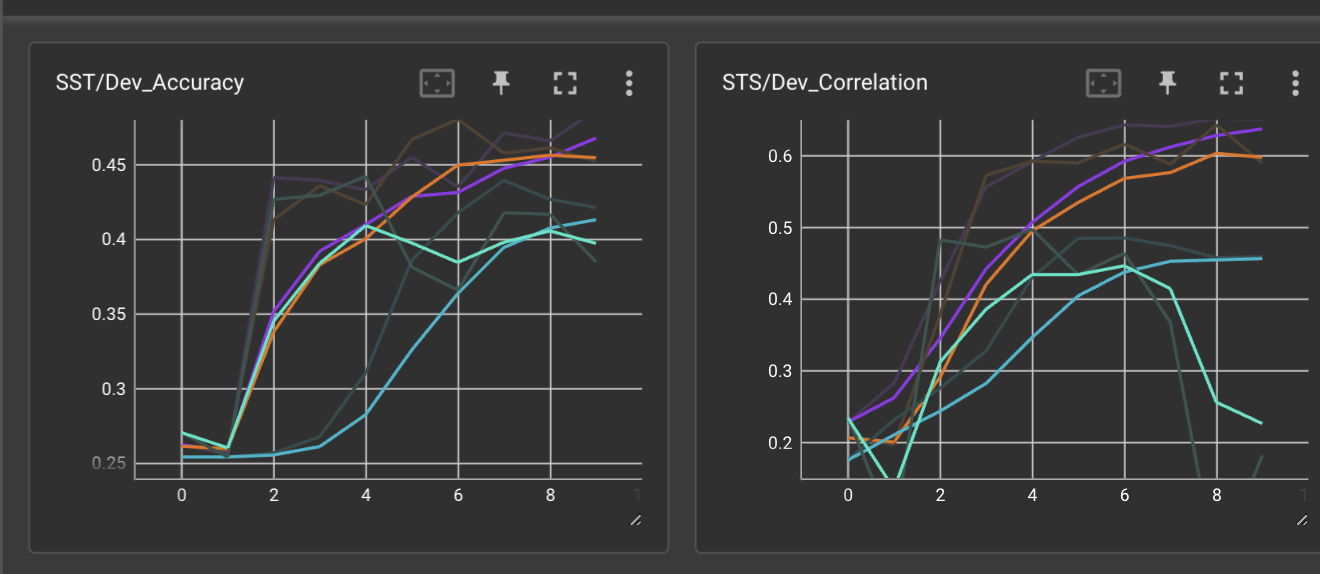}
    \caption{Impact of cyclic learning rate and EMA on performance. Medium Blue is EMA with decay 0.999, Light Blue is EMA with decay 0.5, reference curves are traditional training with cyclic LR scheduling.}
    \label{fig:cyclic_lr_and_ema}
\end{figure}

Unfortunately, our EMA results were very poor. We propose that this is likely a result of (1) the relatively few epochs we train for and (2) our non-traditional implementation choices. With a traditional decay value of $0.999$, we retain far too much of our initial (and low performing) weights due to the very few epochs we train for (see medium blue curve below), and with a lower decay of $0.5$ we see instability (see light blue curve below), especially as training progresses (\autoref{fig:cyclic_lr_and_ema}). Furthermore, our use of extremely short cycles ($2$ epochs), cyclical learning rate with \texttt{Adamax} and EMA (rather than \texttt{SGD} and SWA) are non-traditional and seem to significantly detract from model training performance. 

\clearpage

\section{Custom Loss Functions}
\label{sec:customloss}

\subsection{Optimal Loss Functions}

\[
L_{\text{CE}}(p, y) = -\sum_{i=1}^{C} y_i \log \left( \frac{e^{p_i}}{\sum_{j=1}^{C} e^{p_j}} \right)
\]
where \( C \) is the number of classes, \( p \) are the predicted logits, and \( y \) is the one-hot encoded true label vector.

\[
L_{\text{BCE}}(p, y) = -\frac{1}{N} \sum_{i=1}^{N} \left[ y_i \log(\sigma(p_i)) + (1 - y_i) \log(1 - \sigma(p_i)) \right]
\]
where \( \sigma(p_i) = \frac{1}{1 + e^{-p_i}} \) denotes the sigmoid function, \( p \) represents the logits from the model, \( y \) is the binary label, and \( N \) is the number of samples.

\[
L_{\text{Pearson}}(p, y) = 1 - \frac{\sum_{i=1}^{n} (p_i - \bar{p})(y_i - \bar{y})}{\sqrt{\sum_{i=1}^{n} (p_i - \bar{p})^2} \sqrt{\sum_{i=1}^{n} (y_i - \bar{y})^2}}
\]
where \( \bar{p} \) and \( \bar{y} \) are the means of the predicted and true values, respectively.

The optimization objective is to find the parameter set \( \theta = \{\theta_{\text{SST}}, \theta_{\text{Para}}, \theta_{\text{STS}}, \Omega\} \) that minimizes the total loss \( L \):
\[
\theta^* = \arg\min_{\theta} L(\theta)
\]

\subsection{Other Loss Function Experiments}

%
%
%
%
%
%

\subsection*{Kullback-Leibler Divergence Loss (KL)}
This loss measures how one probability distribution diverges from a second, expected probability distribution:

\[
L = \sum_{i=1}^n y_i \cdot (\log(y_i) - \mathbf{z}_i)
\]

where \( y_i \) represents the target probability for the \(i\)-th instance, and \( \mathbf{z}_i \) denotes the predicted log-probabilities for the \(i\)-th instance.

\subsection*{Mean Squared Error Loss (MSE)}
Typically used for regression:
\[
L = \frac{1}{n} \sum_{i=1}^n (\hat{y}_i - y_i)^2
\]

where \( \hat{y}_i \) represents the predicted value for the \(i\)-th instance and \( y_i \) denotes the actual target value for the \(i\)-th instance.

\subsection*{Log-Cosh Loss}
Log-cosh is a smooth loss that behaves much like the mean squared error but does not yield gradients as steep as MSE:
\[
L = \frac{1}{n} \sum_{i=1}^n \log(\cosh(\hat{y}_i - y_i))
\]

where $\hat{y}_i$represents the predicted value for the ith instance. $y_i$ denotes the actual target value for the ith instance.

\subsection*{BCE/MSE Mixed Loss}
A combination of BCE and MSE, useful for tasks that benefit from both probabilistic and regression-based approaches. We tested this mixed loss on both the SST and STS tasks, with some improvements in early epochs, but ultimately worse convergence. 

\[
L = \lambda_{BCE} \cdot L_{BCE} + \lambda_{MSE} \cdot L_{MSE}
\]

\subsection*{Ordinal Cross Entropy Loss}
This loss function is used for ordinal regression. It involves calculating the cumulative probabilities for each class and then applying the cross entropy loss to these cumulative probabilities. The loss is defined as:
\[
L = -\sum_{i=1}^n \left[ \text{one\_hot}(y_i, k) \cdot \log(\text{cumulative\_probs}_i) + (1 - \text{one\_hot}(y_i, k)) \cdot \log(1 - \text{cumulative\_probs}_i) \right]
\]
where \(\text{one\_hot}(y_i, k)\) is the one-hot encoded vector of the target class label \(y_i\) with \(k\) classes. The cumulative probabilities \(\text{cumulative\_probs}_i\) for a predicted class label \(j\) are calculated as:
\[
\text{cumulative\_probs}_i(j) = \sum_{l=1}^j \text{softmax}(\mathbf{z}_i)(l)
\]
where \(\mathbf{z}_i\) are the logits predicted by the model for the \(i\)-th instance, and \(\text{softmax}(\mathbf{z}_i)(l) = \frac{e^{z_{il}}}{\sum_{m=1}^k e^{z_{im}}}\) is the softmax function applied to these logits.

\subsection*{Contrastive Loss}
Often used in learning embeddings or in siamese networks, contrastive loss helps to ensure that similar items are closer to each other than they are to dissimilar items. The loss is defined as:
\[
L = \sum_{i=1}^n \left[(1 - y_i) \frac{1}{2} (\text{dist}_i)^2 + y_i \frac{1}{2} \max(0, \text{margin} - \text{dist}_i)^2\right]
\]
where \(y_i\) is the binary label indicating whether the pair of instances are similar (1) or dissimilar (0), \(\text{dist}_i\) is the Euclidean distance between the embeddings of the pair, computed as:
\[
\text{dist}_i = \sqrt{\sum_{j=1}^d (x_{i1j} - x_{i2j})^2}
\]
where \(x_{i1j}\) and \(x_{i2j}\) represent the \(j\)-th dimension of the embeddings for the first and second instance of the \(i\)-th pair, respectively, and \(d\) is the dimensionality of the embeddings. The \(\text{margin}\) is a predefined threshold that defines how far dissimilar pairs should be pushed apart.

\begin{lstlisting}[language=Python, caption=Python Code for Various Loss Functions and Training Loops]
def pearson_correlation_loss(preds, target):
    vx = preds - torch.mean(preds)
    vy = target - torch.mean(target)
    cost = torch.sum(vx * vy) / (torch.sqrt(torch.sum(vx ** 2)) * torch.sqrt(torch.sum(vy ** 2)))
    return 1 - cost  # return 1 - correlation to make it a loss

def log_cosh_loss(preds, target):
    def log_cosh(x):
        return x + torch.nn.functional.softplus(-2. * x) - torch.log(torch.tensor(2.0))
    return torch.mean(log_cosh(preds - target))

class OrdinalCrossEntropyLoss(nn.Module):
    def __init__(self):
        super(OrdinalCrossEntropyLoss, self).__init__()

    def forward(self, logits, targets):
        targets_one_hot = F.one_hot(targets, num_classes=logits.size(1)).float()
        cumulative_probs = torch.cumsum(F.softmax(logits, dim=1), dim=1)
        loss = - torch.sum(targets_one_hot * torch.log(cumulative_probs) + (1 - targets_one_hot) * torch.log(1 - cumulative_probs))
        return loss 
\end{lstlisting}

In the training loop for SST,  we tried CE, BCE, KL loss (which can be equivalent to CE but has different convergence properties which may make it less favorable), ordinal CE, and finally a "mixed" version which was a hybrid between  MSE and BCE. We found that the mixed loss led to the best performance for SST.

\begin{lstlisting}
# ======= Various Criterion Options ==========
criterion_ce = nn.CrossEntropyLoss(reduction='mean')
criterion_bce = nn.BCEWithLogitsLoss(reduction='mean')
criterion_kl = nn.KLDivLoss(reduction='batchmean')
criterion_ord = OrdinalCrossEntropyLoss()

if criterion == criterion_ce:
    loss = criterion(logits, b_labels.view(-1)) 
elif criterion == criterion_bce:
    labels_one_hot = F.one_hot(b_labels, num_classes=5).float()
    loss = criterion(logits, labels_one_hot)
elif criterion == criterion_kl:
    labels_one_hot = F.one_hot(b_labels, num_classes=5).float()
    log_probs = F.log_softmax(logits, dim=-1)
    loss = criterion(log_probs, labels_one_hot) 
elif criterion == "mixed": 
    lambda_bce = 1
    lambda_mse = 0.1
    labels_one_hot = F.one_hot(b_labels, num_classes=5).float()
    loss_bce = nn.BCEWithLogitsLoss()(logits, labels_one_hot)
    probabilities = F.softmax(logits, dim=1)
    scores = torch.tensor([1, 2, 3, 4, 5], dtype=torch.float32, 
        device=device).unsqueeze(0)
    expected_scores = torch.matmul(probabilities, scores.T).squeeze(1)
    loss_mse = nn.MSELoss()(expected_scores, b_labels.float())
    loss = lambda_bce * loss_bce + lambda_mse * loss_mse
\end{lstlisting}

In the training loop for STS, we tried MSE, Pearson, constrastive loss, and a hybrid between MSE and Pearson. We found that Pearson loss works best. Contrastive loss led to a slight decrease in performance. This may be because contrastive loss is fundamentally binary, optimizing to categorize pairs as either similar or dissimilar without accounting for degrees of similarity (which is what the STS task includes). The dichotomous nature of contrastive loss doesn't support the precise quantification of semantic closeness required by STS.

\begin{lstlisting}
criterion_mse = nn.MSELoss()  # MSE loss since labels are 0-5
    criterion_pearson = pearson_correlation_loss

        if criterion in [criterion_mse, criterion_pearson]:
            loss = criterion(logits, b_labels) 
        elif criterion == "mixed":
            lambda_mse = 0.5 
            lambda_pears = 1.0
            mse_loss = criterion_mse(logits, b_labels) 
            pears_loss = pearson_correlation_loss(logits, b_labels)
            loss = lambda_mse * mse_loss + lambda_pears * pears_loss
\end{lstlisting}

\clearpage

\section{Chronological Performance Experiments}
\label{sec:chron_exp}

Below we detail the chronological and qualitative experiments performed towards deriving our strongest model architectures and training routines. In early experiments, we worked iteratively and used learning curves obtained via TensorBoard \cite{tensorflow2015tensorboard} as our measure of model training stability and performance. We performed all early experiments using only SST and STS training, due to the massive compute requirements for training the Quora paraphrase task. In occasional interleaved full-training experiments, we found that bench-marking optimal architecture and learning rates using these two tasks served as a fairly reliable proxy full-training performance. Experiments were run for $10$ epochs with an initial learning rate of $2e-4$ and multiplicative learning rate decay of $0.5^{\text{epoch}} \times LR$ unless indicated otherwise. 

\begin{figure}[!ht]
    \centering
    \includegraphics[width=0.5\linewidth]{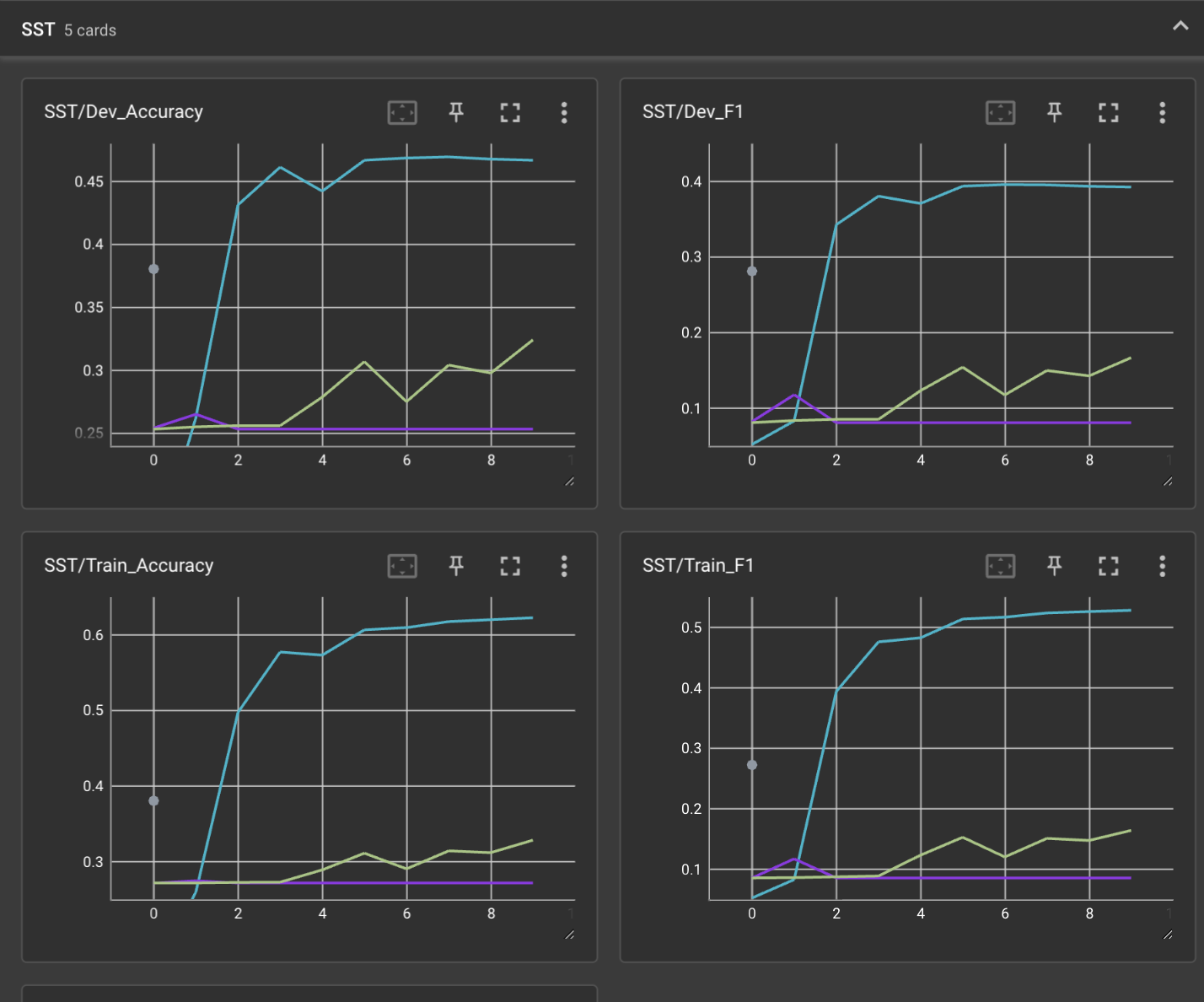}
    \caption{Performance comparison of SGD and Adam variants.}
    \label{fig:adam_vs_sgd}
\end{figure}

First, we tested several different optimizers. SGD (green and purple below) performed much worse than \texttt{Adam} variants (light blue), as shown in \autoref{fig:adam_vs_sgd}.

\texttt{Adamax}, which has been proposed as a suitable optimizer for embeddings, performed better than vanilla \texttt{Adam} and \texttt{SGD} tests on \texttt{SST} and \texttt{STS}. Adding a shared Linear Similarity Feature Extraction layer shared between the Quora and STS classifiers, after the BERT backbone, enabled more improvement (see \autoref{fig:adamax_better}). We hypothesized, and later confirmed, that this shared sub-layer would further improve performance when training for \texttt{Quora}, since \texttt{Quora} and \texttt{STS} are both similarity-based tasks. The shared Similarity Feature Extraction layer might enable these two tasks to learn more from each other even with a frozen backbone. 

Result (no Quora training):
\begin{enumerate}
    \item \textbf{Dev sentiment accuracy:} 0.520
    \item \textbf{Dev paraphrase accuracy:} 0.368
    \item \textbf{Dev STS correlation:} 0.611
\end{enumerate}

\begin{figure}[!ht]
    \centering
    \includegraphics[width=0.5\linewidth]{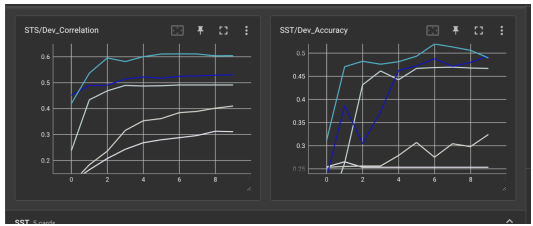}
    \caption{Adamax (dark blue) yields smoother performance increase and better convergence as opposed to Adam and SGD (grey). Adamax + Linear Similarity Feature Extraction layer between minBERT and Quora/STS tasks further improves performance (light blue)}
    \label{fig:adamax_better}
\end{figure}

\begin{figure}[!ht]
    \centering
    \includegraphics[width=0.7\linewidth]{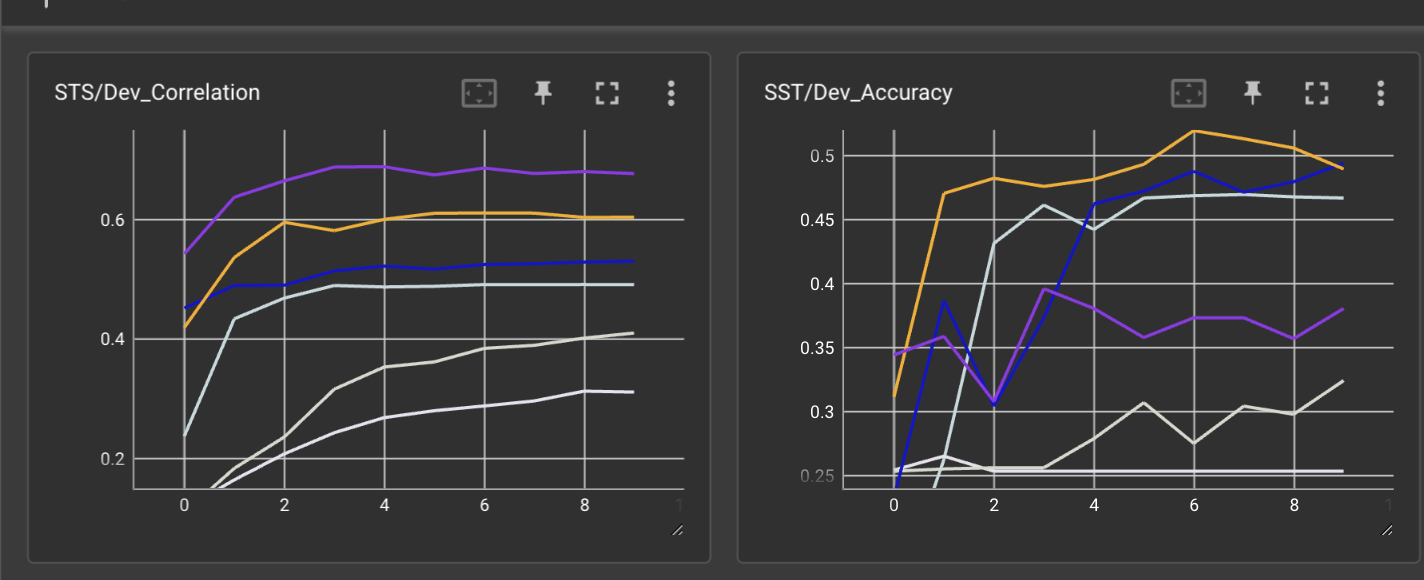}
    \caption{Effect of Pearson Correlation Loss Optimizer (Purple) on STS and SST performance.}
    \label{fig:pearson_loss_improve_sts}
\end{figure}

We then implemented a number custom loss functions, including a Pearson Correlation Loss designed to directly optimize STS for Pearson Correlation (see \autoref{sec:customloss}). The Pearson Correlation Loss Optimizer improved \texttt{STS} performance, but somehow decreased \texttt{SST} performance even though it wasn't applied to \texttt{SST}, as shown in \autoref{fig:pearson_loss_improve_sts}. This impact on SST performance was later resolved by the "Duality of Man" architecture (see \autoref{sec:duality_of_man}). 

Pearson Correlation Loss result (no Quora training):
\begin{enumerate}
    \item \textbf{Dev sentiment accuracy:} 0.396
    \item \textbf{Dev paraphrase accuracy:} 0.368
    \item \textbf{Dev STS correlation:} 0.688
\end{enumerate}

\begin{figure}[!ht]
    \centering
    \includegraphics[width=0.7\linewidth]{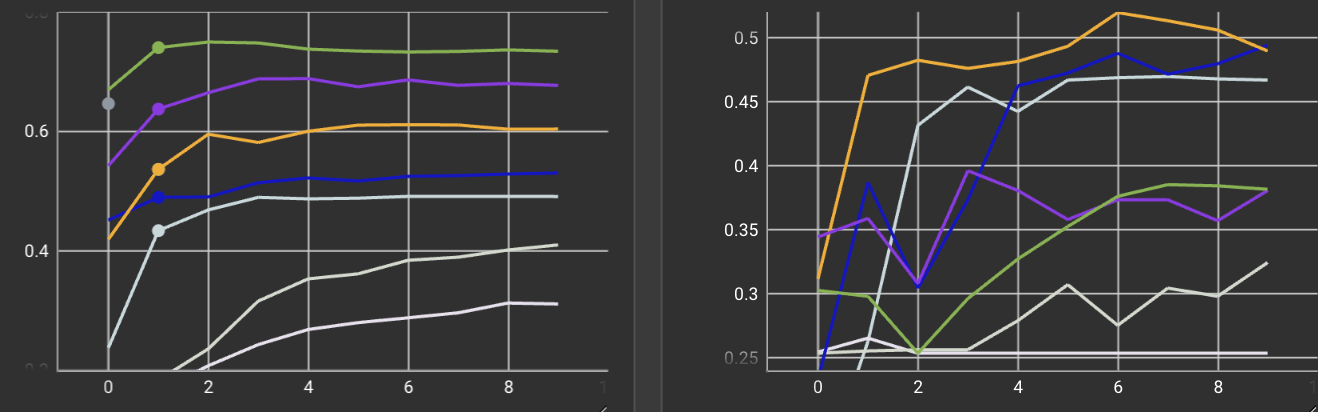}
    \caption{NonLinear Similarity Feature Layer improves STS (light green).}
    \label{fig:shared_nonlinear_sublayer_improves}
\end{figure}

We then upgraded the Similarity Feature Extraction layer to be non-linear instead of linear and noted further STS performance improvements, as shown in \autoref{fig:shared_nonlinear_sublayer_improves}.

Result (no Quora training):
\begin{enumerate}
    \item \textbf{Dev sentiment accuracy:} 0.384
    \item \textbf{Dev paraphrase accuracy:} 0.632
    \item \textbf{Dev STS correlation:} 0.737
\end{enumerate}

\begin{figure}[!ht]
    \centering
    \includegraphics[width=0.5\linewidth]{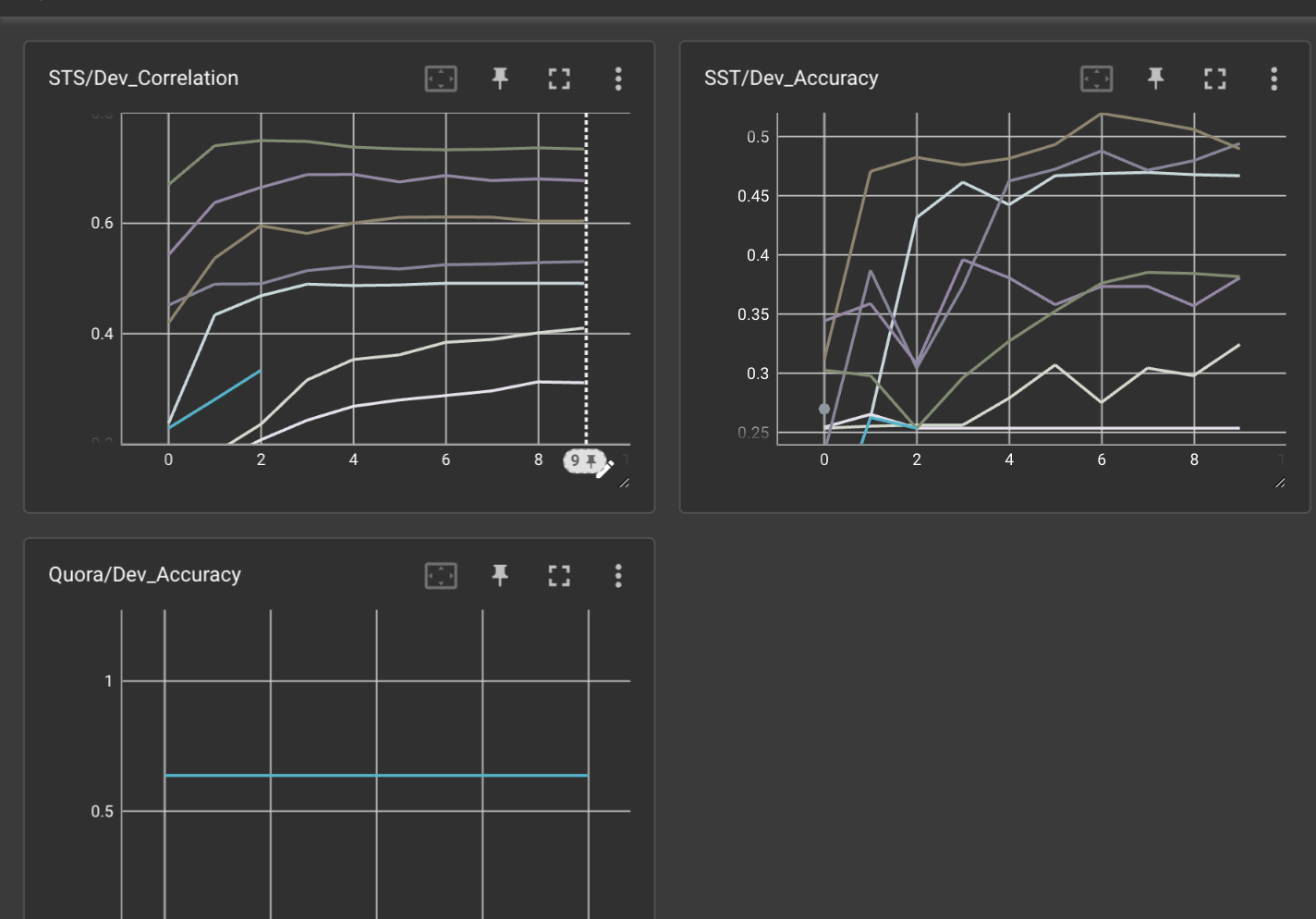}
    \caption{Performance impact of using a small portion of Quora data for training and benchmarking.}
    \label{fig:rough_dirty_quora_eval_is_bad}
\end{figure}

\clearpage
Towards gaining a rough-and-dirty approximation of Quora performance, we then implemented partial Quora training whereby we would randomly select approximately 5\% (200 batches at batch size 96) of \texttt{Quora} data to get a rough and dirty train and benchmark. Unfortunately, this approximation did not work. Performance definitely decreased on the other tasks, as shown in \autoref{fig:rough_dirty_quora_eval_is_bad}. Notably, with full Quora training experiments, we did not see the same performance decrease. 

\begin{figure}[!ht]
    \centering
    \includegraphics[width=0.5\linewidth]{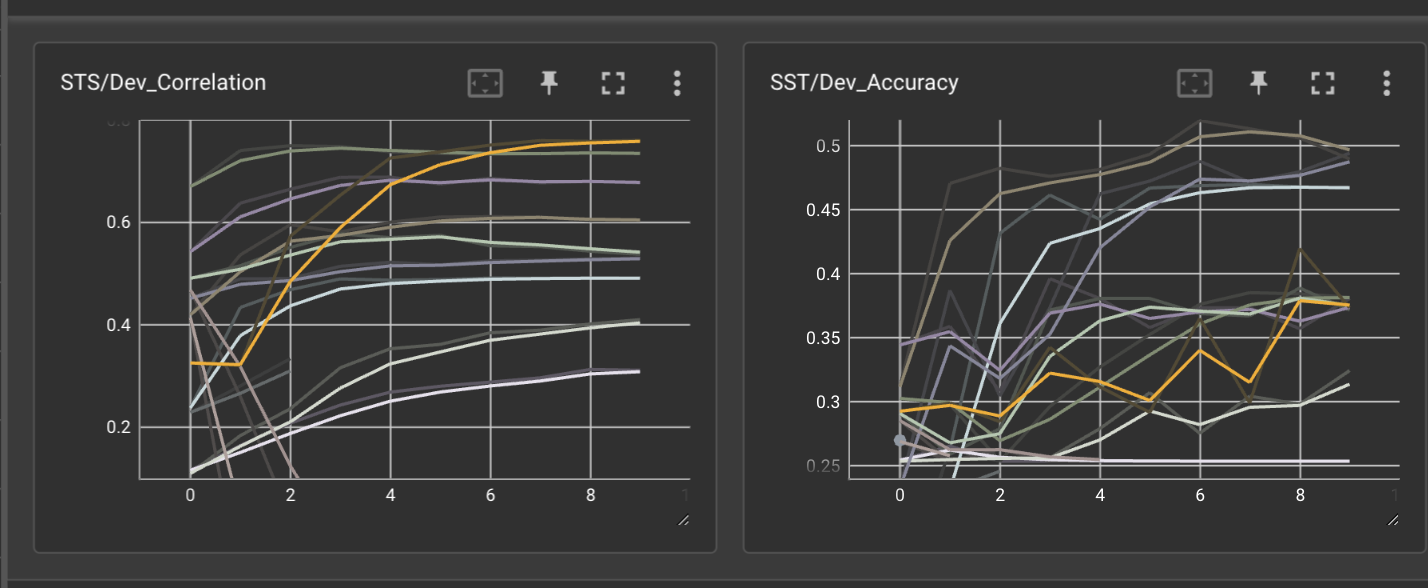}
    \caption{Iterative Unfreezing (Orange) benefits STS convergence. SST suffers instability.}
    \label{fig:iterative_unfreeze_helps_sts}
\end{figure}

We then implemented iterative unfreezing (see \autoref{sec:unfreezing} for details). Iterative unfreezing worked well for \texttt{STS}. However, we noticed some continued instability on \texttt{SST} with an initial learning rate of \texttt{2e-4}, as shown in \autoref{fig:iterative_unfreeze_helps_sts}.

To explore the SST learning instability, we modified initial learning rate, and noticed (unexpectedly, given the SST instability) that faster learning rates were very important to \texttt{SST}, but detrimental to \texttt{STS}. We also found that larger regularization (\texttt{weight\_decay}) hurt \texttt{SST}. See \autoref{fig:sst_faster_sts_slower_lr} below.

\begin{figure}[!ht]
    \centering
    \includegraphics[width=0.5\linewidth]{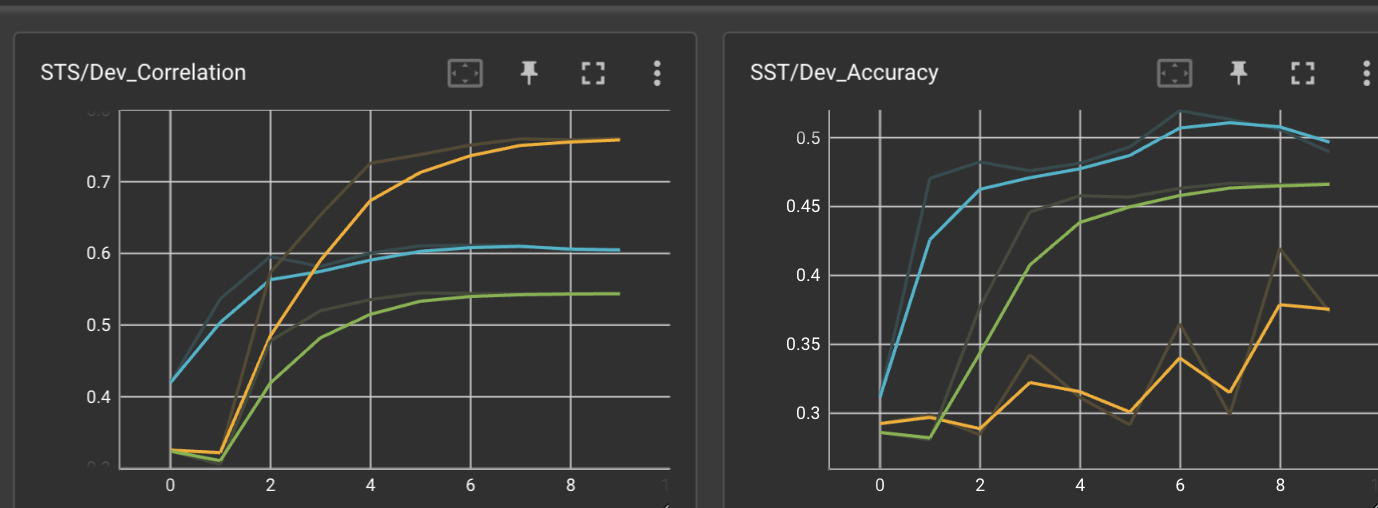}
    \caption{Orange (low learning rate) benefits STS performance while green (high learning rate) benefits SST performance. Large regularization (weight decay of $1e-3$) decreases performance on both SST and STS.}
    \label{fig:sst_faster_sts_slower_lr}
\end{figure}

Towards optimizing learning rate for each sub-task, we then introduced 3 \textit{separate} \texttt{Adamax} optimizers and 3 \textit{separate} learning rate schedulers towards obtaining more degrees of freedom on training optimization for each task. In the first experiment, we multiply \texttt{SST} by \texttt{10} (purple). This did help \texttt{SST} do well and stabilized the learning curves, but \texttt{STS} performance decreased, possibly due to being overwhelmed by \texttt{STS}. This finding that increased sub-task learning rate within a reasonable range ($\approx 1e-4$ to $2e-4$) leads to increased bias towards learning that task held for future experiments as well. See \autoref{fig:sst_10x_lr} below.

\begin{figure}[!ht]
    \centering
    \includegraphics[width=0.5\linewidth]{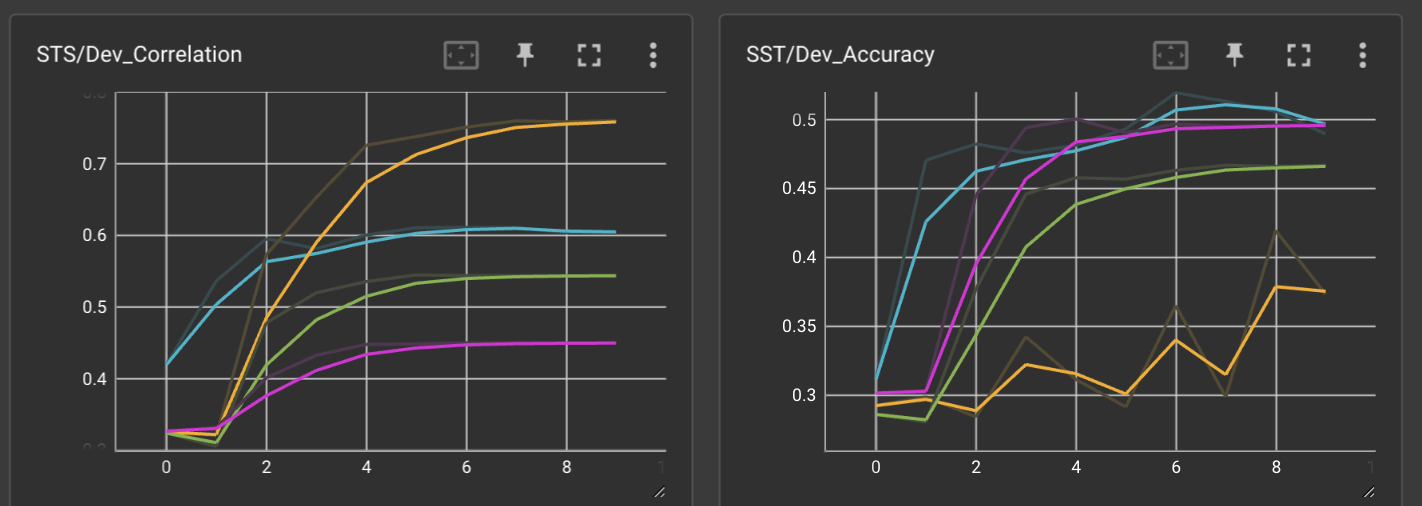}
    \caption{Effect of increasing the initial learning rate for SST by \texttt{10} on training stability and performance.}
    \label{fig:sst_10x_lr}
\end{figure}

To alleviate the sensitivity in balancing SST and STS performance, we separated the minBERT backbone into two separate "Sent" (\texttt{SST}) and "Sim" (\texttt{Quora} + \texttt{STS}) backbones, with moderate performance improvements (see \autoref{sec:duality_of_man} for more dualities on this architecture, dubbed "Duality of Man" for the two backbones). This architecture prevented \texttt{SST} from blowing up the \texttt{STS} performance when increasing the learning rate. Furthermore, introducing some \texttt{nltk} preprocessing (lemmatization, stopword removal, special character cleanup) to the input data for all three tasks yielded minor gains. See \autoref{fig:dark_arts_backbone} below.

Duality of Man + nltk Initial Result (No Quora training):
\begin{enumerate}
    \item \textbf{Dev sentiment accuracy:} 0.510
    \item \textbf{Dev paraphrase accuracy:} 0.368
    \item \textbf{Dev STS correlation:} 0.760
\end{enumerate}

\begin{figure}[!ht]
    \centering
    \includegraphics[width=0.5\linewidth]{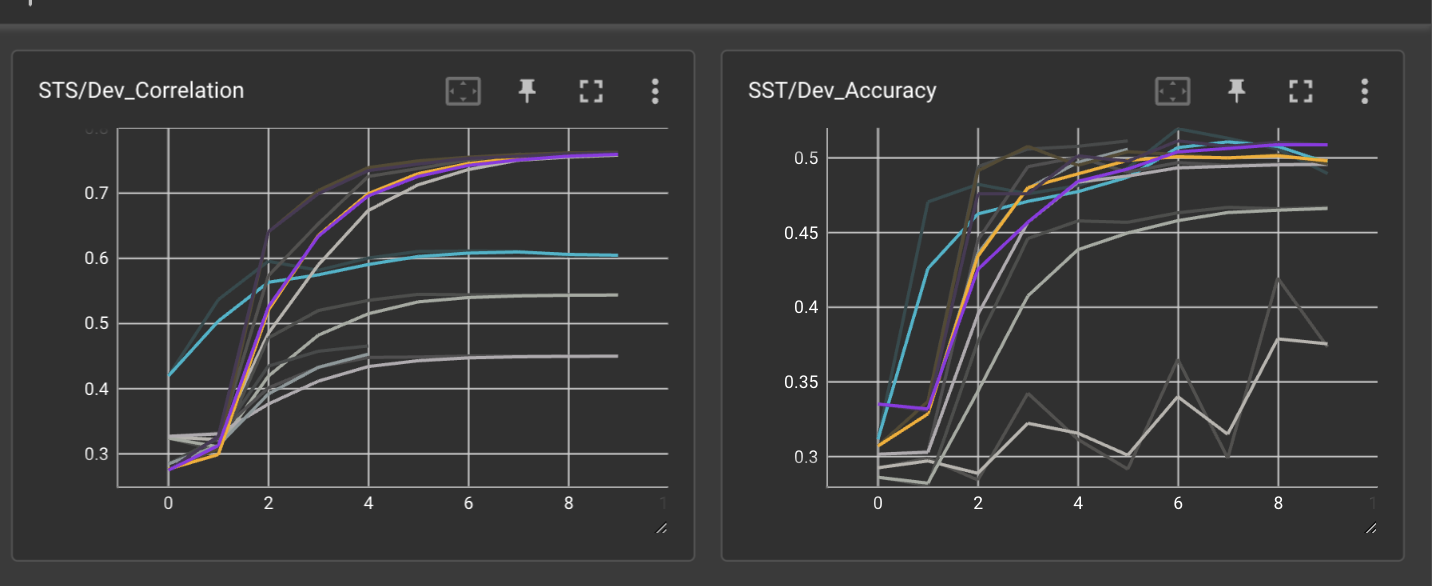}
    \caption{Effect of separating \texttt{minBERT} backbones (orange) and introducing \texttt{nltk} preprocessing (purple) on performance.}
    \label{fig:dark_arts_backbone}
\end{figure}

\begin{figure}[!ht]
    \centering
    \includegraphics[width=0.5\linewidth]{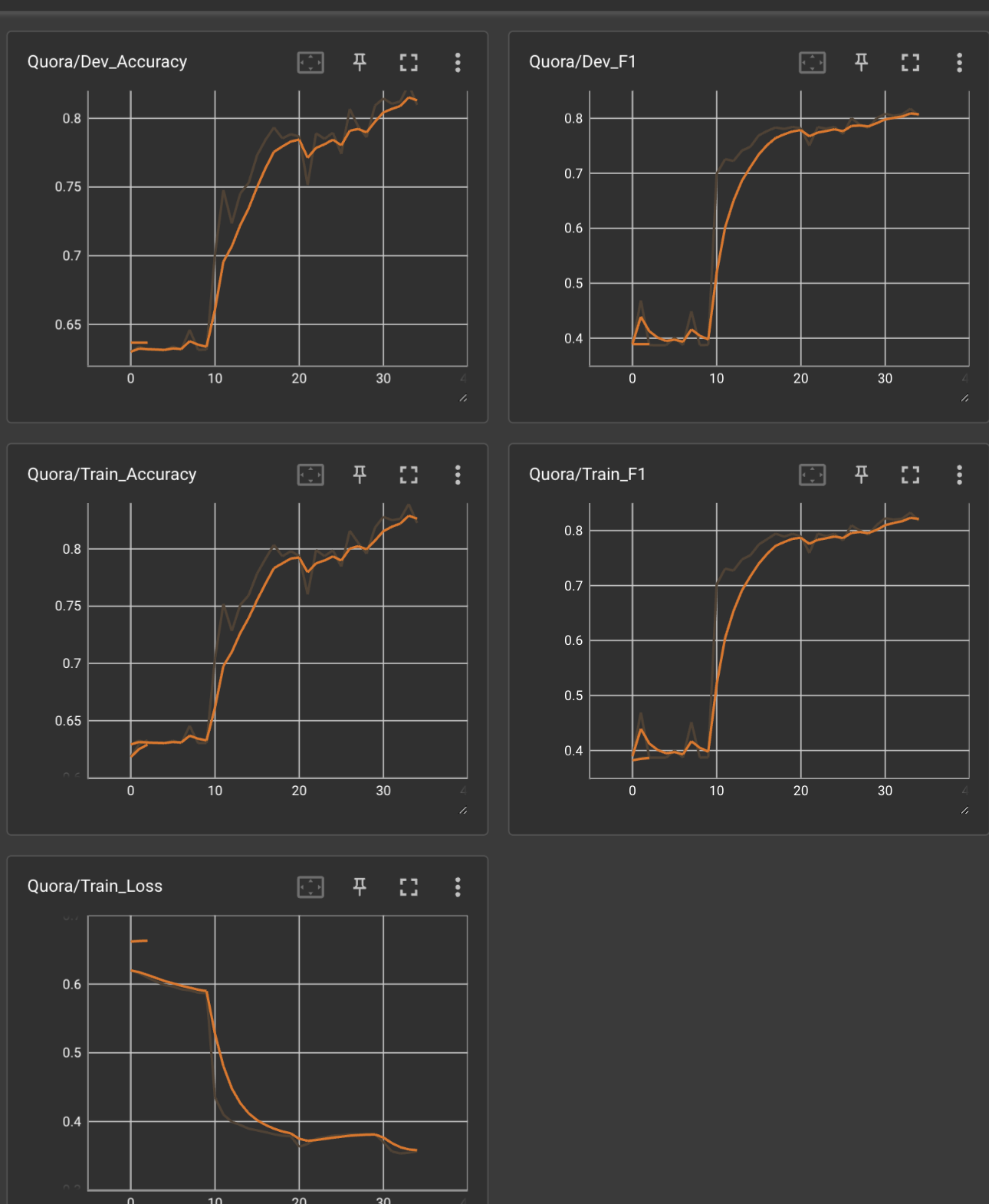}
    \caption{Performance improvement of the Quora dataset after unfreezing the backbone.}
    \label{fig:35_epochs_quora}
\end{figure}

We ran a massive 50 epoch (10 epoch cycles for each unfreezing of \texttt{1/4} of the backbone) experiment for 35 epochs overnight, and noticed that the \texttt{Quora} dataset did not appreciably improve in performance until the backbone was somewhat unfrozen (epoch 10 here), as shown in \autoref{fig:35_epochs_quora}. As such, we implemented a dynamic Quora training option whereby Quora training does not begin until a later stage of unfreezing (though we did not experiment with this much).

\clearpage 

\begin{figure}[!ht]
    \centering
    \includegraphics[width=0.5\linewidth]{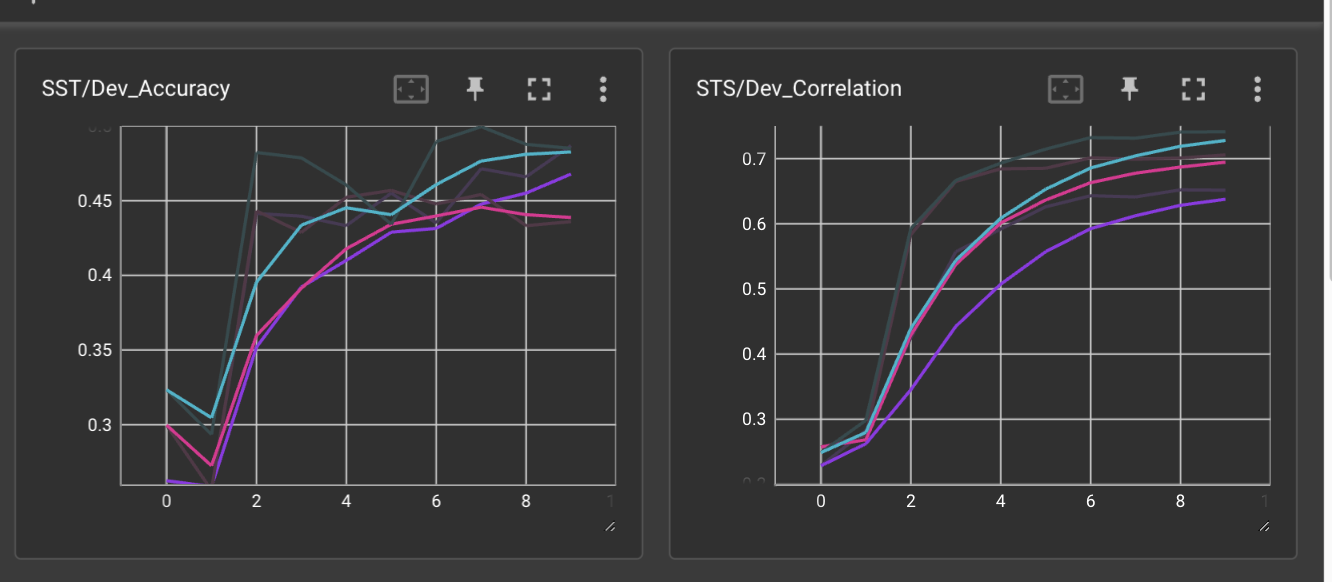}
    \caption{Impact of lower learning rate and \texttt{lambda = 0.7**epoch} scheduler on performance (purple).}
    \label{fig:1e_4_lr_worse_sts_initially}
\end{figure}

Our next experiment used a lower initial learning rate (\texttt{1e-4}) with a shallower learning rate decay ($0.7^{\text{epoch}} * LR$). This led to worse performance. In particular, \texttt{STS} suffered, as shown in \autoref{fig:1e_4_lr_worse_sts_initially}.

Result:
\begin{enumerate}
    \item \textbf{Sentiment classification accuracy:} 0.487
    \item \textbf{Paraphrase detection accuracy:} 0.368
    \item \textbf{Semantic Textual Similarity correlation:} 0.651
\end{enumerate}

\begin{figure}[!ht]
    \centering
    \includegraphics[width=0.5\linewidth]{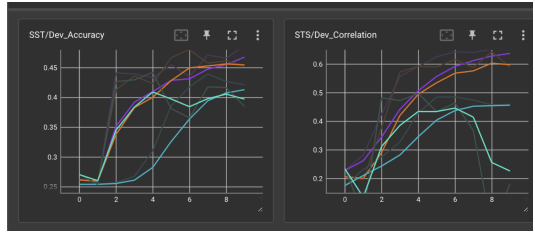}
    \caption{Impact of cyclic learning rate and EMA on performance.}
    \label{fig:cyclic_lr_and_ema2}
\end{figure}

We then tried a triangular cyclical learning rate schedule from $1e-4$ to $2e-4$, seeing very little difference. With the cylical learning rate scheduler still in place, we experiment with an exponential moving average (EMA) wrapper for our model, which decreased performance, especially in later epochs. See \autoref{sec:ema} for details on the EMA experiment. The learning curve is also below (\autoref{fig:cyclic_lr_and_ema2}).

\begin{figure}[!ht]
    \centering
    \includegraphics[width=0.5\linewidth]{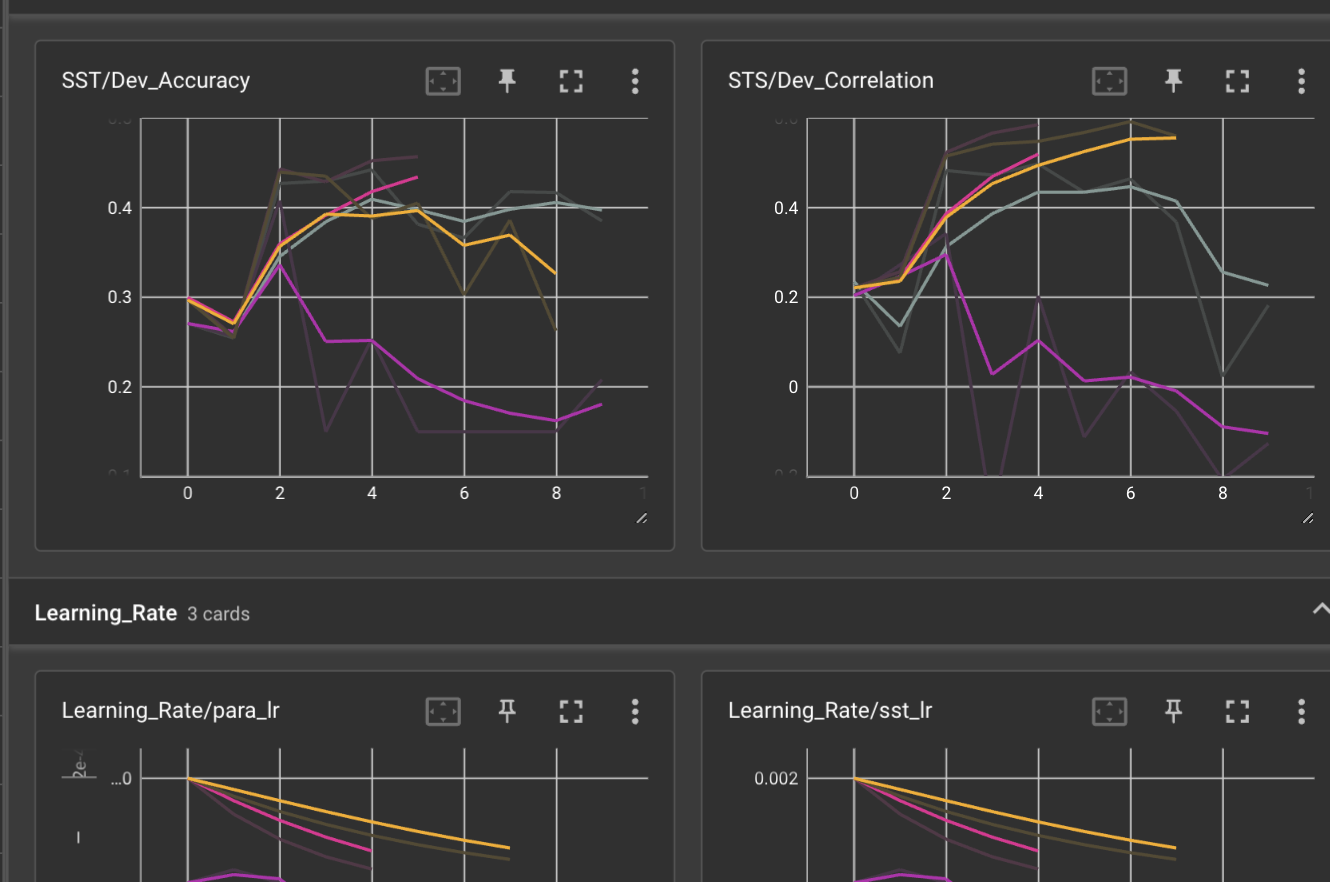}
    \caption{Comparison of Lambda LR schedulers: \texttt{0.7} (pink) vs \texttt{0.85} (orange).}
    \label{fig:0.7_0.85_lambda_lr_exps}
\end{figure}

We then scrapped EMA, and continue exploring optimal multiplicative learning rate scheduling, noting that more aggressive decay (multiplicative factor $0.7$) was better than shallower decay ($0.85$), especially for the SST task. See \autoref{fig:0.7_0.85_lambda_lr_exps}.

\begin{figure}[!ht]
    \centering
    \includegraphics[width=0.5\linewidth]{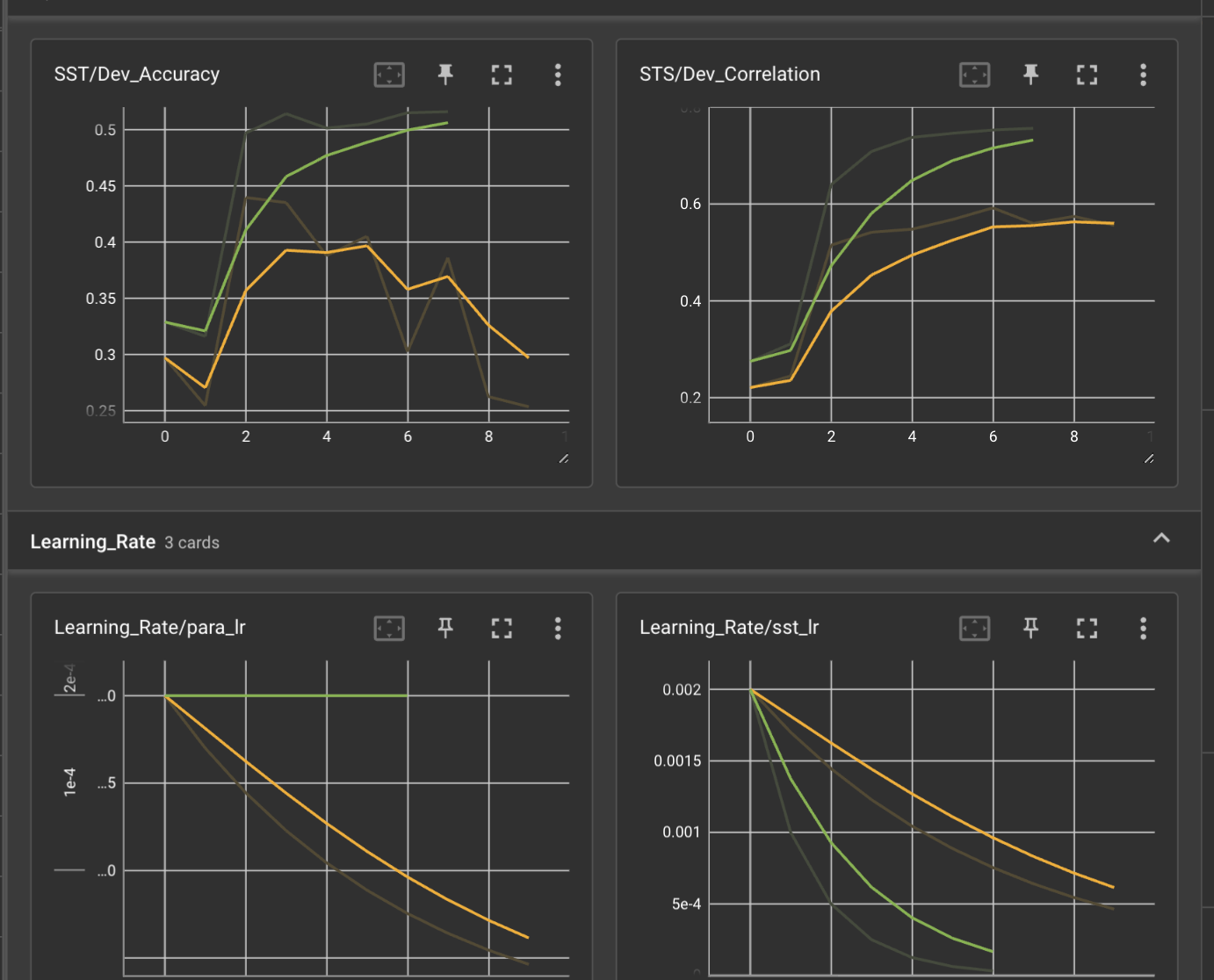}
    \caption{Impact of further reduced learning rate ($0.5$) and less weight decay on learning trajectory.}
    \label{fig:0.5_lr_less_weight_decay}
\end{figure}

Reducing learning rate even more (multiplicative factor of $0.5$) and reducing regularization helped the overall learning trajectory, as shown in \autoref{fig:0.5_lr_less_weight_decay}.

Many further architecture and learning routine experiments were performed, but are not detailed here. See \autoref{sec:best_arch} for our ultimately best architecture and \autoref{sec:ensembling} for our ensembling strategy to achieve maximal multi-task performance. 

\clearpage 

\section{"Duality of Man" Model Variants and Ensemble}
\label{sec:duality_of_man}

\subsection{"Duality of Man" Architecture}

With "Duality of Man" model variants we introduced a separate BERT backbone for the SST task. In experiments, we found that Quora and STS training benefitted one another via shared sub-layers and backbone, likely the result of the inherent similarity between the two tasks. Both the paraphrase detection (Quora) and similarity score (STS) tasks seek to understand how similar two sentences are. On the other hand, the SST task is focused on sentiment, which is largely orthogonal to sentence similarity (especially since we derive sentiment based on the input of one sentence at a time). Attempting to train all three models on the same BERT backbone led to instability in tuning learning rates appropriately so as not to compromise SST performance. By enforcing our prior knowledge that the SST task differs significantly from the other two and partitioning its backbone separately, we saw hugely increased training stability. 

Compared to our best "Yin Yang" model architecture, the earlier "Duality of Man" architecture lacks a few important features. The "Duality of Man" Similarity Feature Extraction sub-layer shared between the paraphrase and STS tasks lacks the unadulterated embeddings of the two sentences. The feature extraction includes only the element-wise product, element-wise difference, and cosine similarity of input sentence embeddings for the Para and STS classifier layers (see below). There are also no BatchNorms in the "Duality of Man" classifier layers. The latter made the more significant difference in boosting "Yin Yang" model variant performance over this earlier "Duality of Man" architecture. 

\begin{lstlisting}
def extract_comparison_features(self,
                           input_ids_1, attention_mask_1,
                           input_ids_2, attention_mask_2):
        """ 
        Given a batch of pairs of sentences, extract comparison features 
        for Quora and STS tasks.
        
        Useful for both Sim tasks.
        """
        
        embeds_1 = self.forward(input_ids_1, attention_mask_1, which_bert = "sim")
        embeds_2 = self.forward(input_ids_2, attention_mask_2, which_bert = "sim")
        cosine_sim = F.cosine_similarity(embeds_1, embeds_2).unsqueeze(-1)
        elem_prods = embeds_1 * embeds_2
        diff = torch.abs(embeds_1 - embeds_2)
        features = torch.cat([diff, elem_prods, cosine_sim], dim=1)
        features = self.comparison_features_fcn(features)
        
        return features
\end{lstlisting}

\textbf{G6 "Duality of Robust Men"} (\texttt{RAdam}, 5x instead of 10x LR for \texttt{SST}), \texttt{1e-4} weight decay for \texttt{para}

\texttt{\$python multitask\_classifier.py --fine-tune-mode iterative --optim RAdam --sst\_lr\_multiplier 5  --para\_weight\_decay 1e-4 --lr 1e-4 --use\_gpu --amp --batch\_size 64 --train\_sst --train\_quora --train\_sts  --clf conv --epochs 10}

\textbf{G6 "Duality of Faster Men"} - same as above, higher LR

\texttt{\$python multitask\_classifier.py --fine-tune-mode iterative --sst\_lr\_multiplier 5 --para\_weight\_decay 1e-4 --lr 2.1e-4 --use\_gpu --amp --batch\_size 64 --train\_sst --train\_quora --train\_sts  --clf conv}

\textbf{G6 "Duality of Slow Men"}

\texttt{\$python multitask\_classifier.py --fine-tune-mode full-model --sst\_lr\_multiplier 5 --para\_weight\_decay 1e-4 --lr 5e-5 --use\_gpu --amp --batch\_size 64 --train\_sst --train\_quora --train\_sts  --clf conv}

\textbf{G6 "Duality of Nonlinear Men"}

\texttt{\$python multitask\_classifier.py --fine-tune-mode iterative --sst\_lr\_multiplier 5 --para\_weight\_decay 1e-4 --lr 1e-4 --use\_gpu --amp --batch\_size 64 --train\_sst --train\_quora --train\_sts  --clf nonlinear}

\textbf{G6 "Mega Duality"} (20 epochs)

\texttt{\$python multitask\_classifier.py --fine-tune-mode iterative --sst\_lr\_multiplier 5 --para\_weight\_decay 1e-4 --lr 1e-4 --use\_gpu --amp --batch\_size 64 --train\_sst --train\_quora --train\_sts  --clf conv --epochs 20}

\subsection{Ensemble Command}

We ensemble all of the Duality of Man variants using the following command. 

\texttt{\$python ensemble.py --filepaths g6-duality-of-man.pt duality-of-fast-man.pt duality-of-long-distance-runners-20.pt duality-of-man-nonlinear.pt duality-of-slow-man.pt duality-of-robust-men.pt}

Result:
\begin{enumerate}
    \item \textbf{Sentiment classification accuracy:} 0.533
    \item \textbf{Paraphrase detection accuracy:} 0.858
    \item \textbf{Semantic Textual Similarity correlation:} 0.853
\end{enumerate}

\clearpage

\section{"Yin Yang" Model Variants and Ensemble}
\label{sec:yinyang}

Towards improving the "Duality of Man" variants architecture, the Yin Yang model architecture (our best multi-task model architecture) includes both pure sentence embeddings and the extracted elementwise product, cosine similarity, and embedding difference features. We also add more BatchNorms to the classifier layers, with the latter providing a significant performance boost. See below for the new shared similarity task feature extraction layer, and the updated classifier layers. 

\begin{lstlisting}

def __init__(self, config):
    ...
    # ===== sentiment =====
    ...
    self.sentiment_conv = nn.Sequential(
        nn.Conv1d(in_channels=1, out_channels=4, kernel_size=3), # note: input needs to be unsqueezed
        nn.BatchNorm1d(4),
        nn.Flatten(),  # Flatten layer to convert (N, 4, D-2) to (N, 4*(D-2))
        nn.Linear(4 * (BERT_HIDDEN_SIZE - 2), BERT_HIDDEN_SIZE // 2),  
        nn.BatchNorm1d(BERT_HIDDEN_SIZE // 2),
        nn.ReLU(),  
        nn.Linear(BERT_HIDDEN_SIZE // 2, N_SENTIMENT_CLASSES)  
    )

    # ====== comparison task shared layer ======
    # learn linear mapping which can be shared among STS and Quora tasks for comparison features
    self.comparison_features_fcn = nn.Sequential(
        nn.Linear(FEATURE_SIZE, FEATURE_PROJ_SIZE),
        nn.BatchNorm1d(FEATURE_PROJ_SIZE),
        nn.ReLU(),
        nn.Linear(FEATURE_PROJ_SIZE, FEATURE_PROJ_SIZE),
        nn.BatchNorm1d(FEATURE_PROJ_SIZE)
    )

    # ===== paraphrase ======
    ...
    self.paraphrase_conv =  nn.Sequential(
        nn.Conv1d(in_channels=1, out_channels=4, kernel_size=3), # note: input needs to be unsqueezed
        nn.BatchNorm1d(4),
        nn.Flatten(),  # Flatten layer to convert (N, 4, D-2) to (N, 4*(D-2))
        nn.Linear(4 * (FEATURE_PROJ_SIZE - 2), BERT_HIDDEN_SIZE // 2),  
        nn.BatchNorm1d(BERT_HIDDEN_SIZE // 2),
        nn.ReLU(),  
        nn.Linear(BERT_HIDDEN_SIZE // 2, 1)  
    )
    ...
    # ===== similarity ========
    ...

    self.similarity_conv = nn.Sequential(
        nn.Conv1d(in_channels=1, out_channels=4, kernel_size=3), # note: input needs to be unsqueezed
        nn.BatchNorm1d(4),
        nn.Flatten(),  # Flatten layer to convert (N, 4, D-2) to (N, 4*(D-2))
        nn.Linear(4 * (FEATURE_PROJ_SIZE - 2), BERT_HIDDEN_SIZE // 2),  
        nn.BatchNorm1d(BERT_HIDDEN_SIZE // 2),
        nn.ReLU(),  
        nn.Linear(BERT_HIDDEN_SIZE // 2, 1)  
    )

...

def extract_comparison_features(self,
                       input_ids_1, attention_mask_1,
                       input_ids_2, attention_mask_2):
    """ 
    Given a batch of pairs of sentences, extract comparison features for Quora and STS tasks.
    
    Useful for both Sim tasks.
    """
    
    embeds_1 = self.forward(input_ids_1, attention_mask_1, which_bert = "sim")
    embeds_2 = self.forward(input_ids_2, attention_mask_2, which_bert = "sim")
    cosine_sim = F.cosine_similarity(embeds_1, embeds_2).unsqueeze(-1)
    elem_prods = embeds_1 * embeds_2
    diff = torch.abs(embeds_1 - embeds_2)
    features = torch.cat([embeds_1, embeds_2, diff, elem_prods, cosine_sim], dim=1)
    features = self.comparison_features_fcn(features)
    
    return features
\end{lstlisting}

\subsection{G6 "Yin Yang"}

\texttt{\$python multitask\_classifier.py --fine-tune-mode iterative --lr 1e-4 --use\_gpu --amp --batch\_size 64 --train\_sst --train\_quora --train\_sts --clf conv --sst\_weight\_decay 8e-3 --para\_weight\_decay 1e-5 --sts\_weight\_decay 9e-3 --lr\_lambda 0.55 --optim Adamax --sst\_lr\_multiplier 7 --para\_lr\_multiplier 1.0 --sts\_lr\_multiplier 3 --epochs 7}

Result:
\begin{enumerate}
    \item \textbf{Dev sentiment accuracy:} 0.516
    \item \textbf{Dev paraphrase accuracy:} 0.851
    \item \textbf{Dev STS correlation:} 0.861
\end{enumerate}

\subsection{G6 "Yang Yin"}

\texttt{\$python multitask\_classifier.py --fine-tune-mode iterative --lr 1e-4 --use\_gpu --amp --batch\_size 64 --train\_sst --train\_quora --train\_sts --clf conv --sst\_weight\_decay 8e-3 --para\_weight\_decay 1e-5 --sts\_weight\_decay 9e-3 --lr\_lambda 0.55 --optim Adamax --sst\_lr\_multiplier 3 --para\_lr\_multiplier 4 --sts\_lr\_multiplier 1 --epochs 7}

Result:
\begin{enumerate}
    \item \textbf{Dev sentiment accuracy:} 0.515
    \item \textbf{Dev paraphrase accuracy:} 0.888
    \item \textbf{Dev STS correlation:} 0.835
\end{enumerate}

\subsection{G6 "Yin Yang Fast"}

\texttt{\$python multitask\_classifier.py --fine-tune-mode iterative --lr 1.5e-4 --use\_gpu --amp --batch\_size 64 --train\_sst --train\_quora --train\_sts --clf conv --sst\_weight\_decay 8e-3 --para\_weight\_decay 1e-5 --sts\_weight\_decay 9e-3 --lr\_lambda 0.55 --optim Adamax --sst\_lr\_multiplier 5 --para\_lr\_multiplier 5 --sts\_lr\_multiplier 2 --epochs 7}

Result:
\begin{enumerate}
    \item \textbf{Dev sentiment accuracy:} 0.513
    \item \textbf{Dev paraphrase accuracy:} 0.887
    \item \textbf{Dev STS correlation:} 0.816
\end{enumerate}

\subsection{G6 "Yin Yang Steady Hand"}

\texttt{\$python multitask\_classifier.py --fine-tune-mode iterative --lr 1e-4 --use\_gpu --amp --batch\_size 64 --train\_sst --train\_quora --train\_sts --clf conv --sst\_weight\_decay 9e-3 --para\_weight\_decay 1e-5 --sts\_weight\_decay 1e-2 --lr\_lambda 0.5 --optim Adamax --sst\_lr\_multiplier 4 --para\_lr\_multiplier 5 --sts\_lr\_multiplier 3 --epochs 7}

Result:
\begin{enumerate}
    \item \textbf{Sentiment classification accuracy:} 0.522
    \item \textbf{Paraphrase detection accuracy:} 0.888
    \item \textbf{Semantic Textual Similarity correlation:} 0.848
\end{enumerate}

\subsection{G6 "Yang Yin The Parabola"}

\texttt{\$python multitask\_classifier.py --fine-tune-mode iterative --lr 1e-4 --use\_gpu --amp --batch\_size 64 --train\_sst --train\_quora --train\_sts --clf conv --sst\_weight\_decay 9e-3 --para\_weight\_decay 1e-5 --sts\_weight\_decay 1e-2 --lr\_lambda 0.5 --optim Adamax --sst\_lr\_multiplier 3 --para\_lr\_multiplier 8 --sts\_lr\_multiplier 1 --epochs 7}

Result:
\begin{enumerate}
    \item \textbf{Sentiment classification accuracy:} 0.515
    \item \textbf{Paraphrase detection accuracy:} 0.892
    \item \textbf{Semantic Textual Similarity correlation:} 0.801
\end{enumerate}

\subsection{G6 "Yin Yang" 2}

Minor adjustments to the original to better time the SST optimal point towards the 7th epoch.

\texttt{\$python multitask\_classifier.py --fine-tune-mode iterative --lr 1e-4 --use\_gpu --amp --batch\_size 64 --train\_sst --train\_quora --train\_sts --clf conv --sst\_weight\_decay 9e-3 --para\_weight\_decay 1e-5 --sts\_weight\_decay 1e-2 --lr\_lambda 0.5 --optim Adamax --sst\_lr\_multiplier 4 --para\_lr\_multiplier 1.0 --sts\_lr\_multiplier 3 --epochs 7}

Result:
\begin{enumerate}
    \item \textbf{Sentiment classification accuracy:} 0.520
    \item \textbf{Paraphrase detection accuracy:} 0.842
    \item \textbf{Semantic Textual Similarity correlation:} 0.858
\end{enumerate}

\subsection{G6 "Yue the Moon Spirit" - "Yin Yang" Ensemble}

We ensemble all of the Yin Yang variants (other than Steady Hand 2,3,4,5,6 which were not yet available) via:

\texttt{\$python ensemble.py --filepaths g6-yin-yang.pt g6-yang-yin.pt g6-yin-yang-fast.pt g6-yin-yang-steady-hand.pt g6-yang-yin-the-parabola.pt g6-yin-yang-2.pt}

Result:
\begin{enumerate}
    \item \textbf{Sentiment classification accuracy:} 0.522
    \item \textbf{Paraphrase detection accuracy:} 0.889
    \item \textbf{Semantic Textual Similarity correlation:} 0.858
\end{enumerate}

Based on the success of the Steady Hand variant, we later made more to add to our final ensembles. 

\subsection{G6 "Yin Yang Steady Hand" 2}

\texttt{\$python multitask\_classifier.py --fine-tune-mode iterative --lr 1e-4 --use\_gpu --amp --batch\_size 64 --train\_sst --train\_quora --train\_sts --clf conv --sst\_weight\_decay 9e-3 --para\_weight\_decay 1e-5 --sts\_weight\_decay 1e-2 --lr\_lambda 0.5 --optim Adamax --sst\_lr\_multiplier 3 --para\_lr\_multiplier 4 --sts\_lr\_multiplier 2 --epochs 7}

Individual results not recorded.

\subsection{G6 "Yin Yang Steady Hand" 3}

\texttt{\$python multitask\_classifier.py --fine-tune-mode iterative --lr 1e-4 --use\_gpu --amp --batch\_size 64 --train\_sst --train\_quora --train\_sts --clf conv --sst\_weight\_decay 9e-3 --para\_weight\_decay 1e-5 --sts\_weight\_decay 1e-2 --lr\_lambda 0.55 --optim Adamax --sst\_lr\_multiplier 4 --para\_lr\_multiplier 5 --sts\_lr\_multiplier 3 --epochs 7}

Individual results not recorded.

\subsection{G6 "Yin Yang Steady Hand" 4}

\texttt{\$python multitask\_classifier.py --fine-tune-mode iterative --lr 1e-4 --use\_gpu --amp --batch\_size 64 --train\_sst --train\_quora --train\_sts --clf conv --sst\_weight\_decay 9e-3 --para\_weight\_decay 1e-5 --sts\_weight\_decay 1e-2 --lr\_lambda 0.6 --optim Adamax --sst\_lr\_multiplier 2 --para\_lr\_multiplier 3 --sts\_lr\_multiplier 2 --epochs 7}

\begin{enumerate}
    \item \textbf{Sentiment classification accuracy:} 0.510
    \item \textbf{Paraphrase detection accuracy:} 0.887
    \item \textbf{Semantic Textual Similarity correlation:} 0.845
\end{enumerate}

\subsection{G6 "Yin Yang Steady Hand" 5}

\texttt{\$python multitask\_classifier.py --fine-tune-mode iterative --lr 1e-4 --use\_gpu --amp --batch\_size 64 --train\_sst --train\_quora --train\_sts --clf conv --sst\_weight\_decay 9e-3 --para\_weight\_decay 1e-5 --sts\_weight\_decay 1e-2 --lr\_lambda 0.5 --optim Adamax --sst\_lr\_multiplier 5 --para\_lr\_multiplier 6 --sts\_lr\_multiplier 4 --epochs 7}

\begin{enumerate}
    \item \textbf{Sentiment classification accuracy:} 0.520
    \item \textbf{Paraphrase detection accuracy:} 0.888
    \item \textbf{Semantic Textual Similarity correlation:} 0.841
\end{enumerate}

\subsection{G6 "Yin Yang Steady Hand" 6}

\texttt{\$python multitask\_classifier.py --fine-tune-mode iterative --lr 1e-4 --use\_gpu --amp --batch\_size 64 --train\_sst --train\_quora --train\_sts --clf conv --sst\_weight\_decay 9e-3 --para\_weight\_decay 1e-5 --sts\_weight\_decay 1e-2 --lr\_lambda 0.5 --optim Adamax --sst\_lr\_multiplier 6 --para\_lr\_multiplier 7 --sts\_lr\_multiplier 5 --epochs 7}

\begin{enumerate}
    \item \textbf{Sentiment classification accuracy:} 0.518
    \item \textbf{Paraphrase detection accuracy:} 0.891
    \item \textbf{Semantic Textual Similarity correlation:} 0.840
\end{enumerate}

\clearpage 

\section{G6 "Run it Back" Model Variant}
\label{sec:runitback_model}

Given the massively larger compute requirements of training full epochs on the Quora dataset, we proposed that we might get easy value with minimal added compute by re-shuffling and re-training the SST and STS tasks multiple times per epoch. As such, we implemented simple training loops to allow multiple rounds of training for each sub-task per-epoch when requested. Unfortunately, this experiment proved largely unsuccessful across all three tasks, with dramatic decrease in paraphrase accuracy (the task we did not multiplicatively train per epoch). The main model architecture here was derived from the "Yin Yang" architecture, with differences limited to training routine.

Command:

\texttt{\$python multitask\_classifier.py --fine-tune-mode iterative --lr 1e-4 --use\_gpu --amp --batch\_size 64 --train\_sst --train\_quora --train\_sts --clf conv --sst\_weight\_decay 9e-3 --para\_weight\_decay 1e-5 --sts\_weight\_decay 1e-2 --lr\_lambda 0.55 --optim Adamax --sst\_lr\_multiplier 4 --para\_lr\_multiplier 1.0 --sts\_lr\_multiplier 3 --epochs 7 --num\_sst\_trains 2 --num\_quora\_trains 1 --num\_sts\_trains 5}

\textbf{Results:}
\begin{enumerate}
    \item \textbf{Dev sentiment accuracy:} 0.494
    \item \textbf{Dev paraphrase accuracy:} 0.653
    \item \textbf{Dev STS correlation:} 0.839
\end{enumerate}

\clearpage

\section{G6 "Interleaver" Model Variants and Ensemble}
\label{sec:interleaver_models}

As opposed to the Duality of Man, Yin Yang, and Run it Back architectures, the Interleaver models derive their name from the fact that we interleave model training at a finer granularity. Whereas all other models are trained by iterating through the full SST Dataloader, then the Quora Dataloader, then the STS Dataloader, this model iterates through SST, Quora, and STS batch by batch, calculating each loss function and backpropagating in turn in a single batch iteration. While we observed inferior performance for the individual models (particularly on SST), the overall ensemble yields surprisingly good results. The main model architecture here was derived from the "Yin Yang" architecture, with differences limited to training routine.

\begin{lstlisting}[language=bash, caption=G6 Interleaver Shell Commands]
# G6 interleaver
$python multitask_classifier.py --fine-tune-mode iterative --lr 1e-4 --use_gpu --amp --batch_size 64 --train_sst --train_quora --train_sts --clf conv --sst_weight_decay 8e-3 --para_weight_decay 1e-5 --sts_weight_decay 9e-3 --lr_lambda 0.55 --optim Adamax --sst_lr_multiplier 4 --para_lr_multiplier 1.0 --sts_lr_multiplier 3 --epochs 7 --num_sst_trains 5 --num_quora_trains 1 --num_sts_trains 5

# G6 interleaver more para
$python multitask_classifier.py --fine-tune-mode iterative --lr 1e-4 --use_gpu --amp --batch_size 64 --train_sst --train_quora --train_sts --clf conv --sst_weight_decay 8e-3 --para_weight_decay 1e-5 --sts_weight_decay 9e-3 --lr_lambda 0.55 --optim Adamax --sst_lr_multiplier 3 --para_lr_multiplier 4 --sts_lr_multiplier 2 --epochs 7 --num_sst_trains 4 --num_quora_trains 1 --num_sts_trains 4

# G6 interleaver fast
$python multitask_classifier.py --fine-tune-mode iterative --lr 1.5e-4 --use_gpu --amp --batch_size 64 --train_sst --train_quora --train_sts --clf conv --sst_weight_decay 8e-3 --para_weight_decay 1e-5 --sts_weight_decay 9e-3 --lr_lambda 0.55 --optim Adamax --sst_lr_multiplier 5 --para_lr_multiplier 5 --sts_lr_multiplier 2 --epochs 7 --num_sst_trains 3 --num_quora_trains 1 --num_sts_trains 3

# G6 interleaver steady hand
$python multitask_classifier.py --fine-tune-mode iterative --lr 1e-4 --use_gpu --amp --batch_size 64 --train_sst --train_quora --train_sts --clf conv --sst_weight_decay 9e-3 --para_weight_decay 1e-5 --sts_weight_decay 1e-2 --lr_lambda 0.5 --optim Adamax --sst_lr_multiplier 4 --para_lr_multiplier 5 --sts_lr_multiplier 3 --epochs 7 --num_sst_trains 5 --num_quora_trains 1 --num_sts_trains 5

# G6 interleaver the parabola
$python multitask_classifier.py --fine-tune-mode iterative --lr 1e-4 --use_gpu --amp --batch_size 64 --train_sst --train_quora --train_sts --clf conv --sst_weight_decay 9e-3 --para_weight_decay 1e-5 --sts_weight_decay 1e-2 --lr_lambda 0.5 --optim Adamax --sst_lr_multiplier 3 --para_lr_multiplier 8 --sts_lr_multiplier 2 --epochs 7 --num_sst_trains 2 --num_quora_trains 2 --num_sts_trains 2

# G6 interleaver 2
$python multitask_classifier.py --fine-tune-mode iterative --lr 1e-4 --use_gpu --amp --batch_size 64 --train_sst --train_quora --train_sts --clf conv --sst_weight_decay 9e-3 --para_weight_decay 1e-5 --sts_weight_decay 1e-2 --lr_lambda 0.5 --optim Adamax --sst_lr_multiplier 4 --para_lr_multiplier 2.0 --sts_lr_multiplier 4 --epochs 7 --num_sst_trains 10 --num_quora_trains 1 --num_sts_trains 10

# ensemble
$python ensemble.py --filepaths g6-interleaver.pt g6-interleaver-more-para.pt g6-interleaver-fast.pt g6-interleaver-steady-hand.pt g6-interleaver-the-parabola.pt g6-interleaver-2.pt
\end{lstlisting}

\subsection*{G6 "Interleaver" Ensemble Results}
Result:
\begin{enumerate}
    \item \textbf{Sentiment classification accuracy:} 0.500
    \item \textbf{Paraphrase detection accuracy:} 0.897
    \item \textbf{Semantic Textual Similarity correlation:} 0.866
\end{enumerate}

\clearpage

\section{Final Ensembles}
\label{sec:final_ensembles}

\subsection{G6 "Ménage à Onze" - Duality of Man + Yin Yang Ensemble}

We ensemble all Yin Yang and Duality of Man variants with:

\texttt{\$python ensemble.py --filepaths g6-yin-yang.pt g6-yang-yin.pt g6-yin-yang-fast.pt g6-yin-yang-steady-hand.pt g6-yang-yin-the-parabola.pt g6-yin-yang-2.pt g6-duality-of-man.pt duality-of-fast-man.pt duality-of-long-distance-runners-20.pt duality-of-man-nonlinear.pt duality-of-slow-man.pt duality-of-robust-men.pt}

Result:
\begin{enumerate}
    \item \textbf{Sentiment classification accuracy:} 0.538
    \item \textbf{Paraphrase detection accuracy:} 0.879
    \item \textbf{Semantic Textual Similarity correlation:} 0.860
\end{enumerate}

\textbf{Other Ensembles Tried:}

\textbf{G6 "Oligarchy" v1 [First CS224N Class Test Leaderboard Submission]}

This model is the same as G6 "Ménage à Onze" except that we now give Yin Yang Steady Hand (our most performant single sub-model) $3$ votes. 

\texttt{\$python ensemble.py --filepaths g6-yin-yang.pt g6-yang-yin.pt g6-yin-yang-fast.pt g6-yin-yang-steady-hand.pt g6-yang-yin-the-parabola.pt g6-yin-yang-2.pt g6-duality-of-man.pt duality-of-fast-man.pt duality-of-long-distance-runners-20.pt duality-of-man-nonlinear.pt duality-of-slow-man.pt duality-of-robust-men.pt g6-yin-yang-steady-hand.pt g6-yin-yang-steady-hand.pt}

\textbf{Dev Result:}
\begin{enumerate}
    \item \textbf{Sentiment classification accuracy:} 0.536
    \item \textbf{Paraphrase detection accuracy:} 0.882
    \item \textbf{Semantic Textual Similarity correlation:} 0.860
\end{enumerate}

\textbf{Test Result: }
\begin{enumerate}
    \item \textbf{SST test accuracy:} 0.551 
    \item \textbf{Paraphrase test accuracy:} 0.881 
    \item \textbf{STS test correlation:} 0.853 
    \item \textbf{Overall test score:} 0.786 
\end{enumerate}

\textbf{G6 "Oligarchy" v2}

This ensemble is the same as Oligarchy v1 except we give Yin Yang, Yang Yin, Yang Yin the Parabola, and Duality of Robust Men models additional votes as well. 

\texttt{\$python ensemble.py --filepaths g6-yin-yang.pt g6-yang-yin.pt g6-yin-yang-fast.pt g6-yin-yang-steady-hand.pt g6-yang-yin-the-parabola.pt g6-yin-yang-2.pt g6-duality-of-man.pt duality-of-fast-man.pt duality-of-long-distance-runners-20.pt duality-of-man-nonlinear.pt duality-of-slow-man.pt duality-of-robust-men.pt g6-yin-yang-steady-hand.pt g6-yin-yang-steady-hand.pt g6-yin-yang.pt g6-yang-yin.pt g6-yang-yin-the-parabola.pt duality-of-robust-men.pt}

Result:
\begin{enumerate}
    \item \textbf{Sentiment classification accuracy:} 0.530
    \item \textbf{Paraphrase detection accuracy:} 0.883
    \item \textbf{Semantic Textual Similarity correlation:} 0.860
\end{enumerate}

\textbf{G6 "Lisan al Gaib"}

This ensemble includes one of (almost) everything, including additional slightly modified versions of Yin Yang Steady Hand. Kitchen-sink style ensemble. It sees across the dunes. 

\texttt{\$python ensemble.py --filepaths g6-yin-yang.pt g6-yang-yin.pt g6-yin-yang-fast.pt g6-yin-yang-steady-hand.pt g6-yang-yin-the-parabola.pt g6-yin-yang-2.pt g6-duality-of-man.pt duality-of-fast-man.pt duality-of-long-distance-runners-20.pt duality-of-man-nonlinear.pt duality-of-slow-man.pt duality-of-robust-men.pt g6-interleaver.pt g6-interleaver-more-para.pt g6-interleaver-fast.pt g6-interleaver-steady-hand.pt g6-interleaver-the-parabola.pt g6-interleaver-2.pt g6-yin-yang-steady-hand-2.pt g6-yin-yang-steady-hand-3.pt g6-yin-yang-steady-hand-4.pt g6-yin-yang-steady-hand-5.pt g6-yin-yang-steady-hand-6.pt}

Result: 
\begin{enumerate}
    \item \textbf{Sentiment classification accuracy:} 0.530
    \item \textbf{Paraphrase detection accuracy:} 0.891
    \item \textbf{Semantic Textual Similarity correlation:} 0.864
\end{enumerate}

\subsection*{G6 "The Sage"}

This ensemble combines primarily known high-quality sub-models, with a handful of weaker performers. It does not perform better than the kitchen sink model, suggesting that simply including more models is more optimal.

\texttt{\$python ensemble.py --filepaths g6-yin-yang.pt g6-yang-yin.pt g6-yin-yang-fast.pt g6-yin-yang-steady-hand.pt g6-yang-yin-the-parabola.pt g6-yin-yang-2.pt g6-duality-of-man.pt duality-of-man-nonlinear.pt duality-of-slow-man.pt duality-of-robust-men.pt g6-interleaver-steady-hand.pt g6-yin-yang-steady-hand-2.pt g6-yin-yang-steady-hand-3.pt g6-yin-yang-steady-hand-4.pt g6-yin-yang-steady-hand-5.pt g6-yin-yang-steady-hand-6.pt}

Result:
\begin{enumerate}
    \item \textbf{Sentiment classification accuracy:} 0.529
    \item \textbf{Paraphrase detection accuracy:} 0.889
    \item \textbf{Semantic Textual Similarity correlation:} 0.860
\end{enumerate}

\clearpage
\section{Efficiency Experiments Details}

\subsection{Grid Search Details}
\label{subsec:grid_search_details}

\begin{table}[h]
\centering
\begin{tabular}{|c|c|}
\hline
\textbf{Parameter} & \textbf{Values} \\
\hline
AMP & On \\
\hline
Learning Rate & $1 \times 10^{-4}$, $5 \times 10^{-5}$, $1 \times 10^{-5}$, $5 \times 10^{-6}$ \\
\hline
Batch Size & 64, 128, 256, 384 \\
\hline
LoRA Rank & 1, 5, 10 \\
\hline
LoRA Mode & 'none', 'all-lin', 'attn', 'all-lin-only', 'attn-only' \\
\hline
DoRA & Off, On \\
\hline
Fine-Tune Mode & 'full-model', 'last-layer' \\
\hline
Epochs & 3 \\
\hline
\end{tabular}
\caption{Grid search parameters for evaluating the impact of LoRA and DoRA.}
\label{tab:params}
\end{table}
\clearpage
\subsection{LoRA and DoRA Over Many Ranks}
\label{subsec:lora_dora_many_r}

This grid search generated a total of $960$ possible combinations. However, only 416 of these combinations are feasible, as certain parameter configurations are logically incompatible (e.g., setting 'LoRA Mode' to 'none' while varying 'LoRA Ranks' is illogical since LoRA is deactivated). Out of these, only 270 could be executed to completion due to memory constraints. For instance, configurations with 'Fine-Tune Mode' set to 'full-model' and 'LoRA Mode' set to 'none' could not support batch sizes above 128. We restricted our training and evaluation to the SST and STS tasks, as the Paraphrasing task required significantly more computational resources.

\begin{figure}[ht]
    \centering
    \includegraphics[scale=0.37]{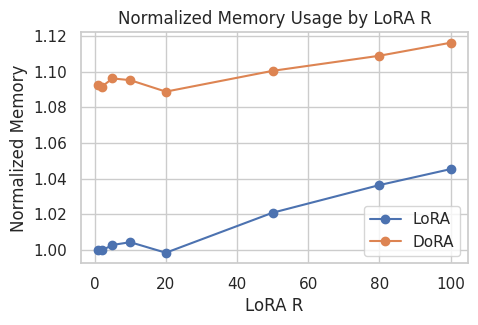}
    \includegraphics[scale=0.37]{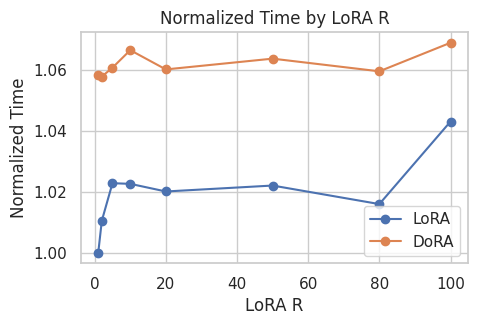}
    \includegraphics[scale=0.37]{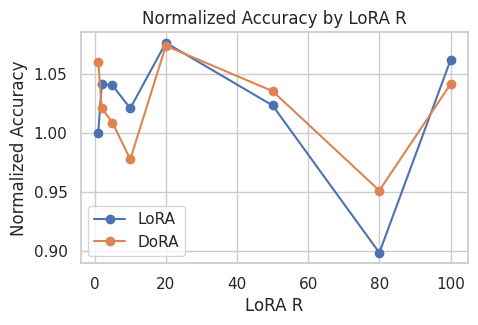}
    \caption{All values normalized to the value for LoRA and Rank 1}
    \label{fig:norm_mem_time_acc_v_r}
\end{figure}

We ran both LoRA and DoRA for ranks 1,2,5,10,20,50,80,100 with the hyperparameters outlined in \autoref{tab:hyperparameters_many_ranks}. The resulting normalized Memory, Time and Accuracy values are presented in \autoref{fig:norm_mem_time_acc_v_r}.

\begin{table}[ht]
\centering
\caption{Hyperparameters}
\label{tab:hyperparameters_many_ranks}
\begin{tabular}{lc}
\toprule
Hyperparameter & Value \\
\midrule
Batch Size & 192 \\
Learning Rate & 0.0002 \\
Epochs & 10 \\
AMP & On \\
Dropout Probability & 0.3 \\
\bottomrule
\end{tabular}
\end{table}

\clearpage
\subsection{Ablation Experiment Details}

\begin{table}[h]
\centering
\begin{tabular}{|c|c|}
\hline
\textbf{Parameter} & \textbf{Values} \\
\hline
AMP & Off, On \\
\hline
Learning Rate & $1 \times 2\times 10^{-4}$ \\
\hline
Batch Size & 96 \\
\hline
LoRA Rank & 1 \\
\hline
LoRA Mode & 'none', 'attn', 'attn-only' \\
\hline
DoRA & Off, On \\
\hline
Fine-Tune Mode & 'full-model', 'last-layer' \\
\hline
Epochs & 10 \\
\hline
\end{tabular}
\caption{Grid search parameters for evaluating the impact of LoRA and DoRA.}
\label{tab:ablation_params}
\end{table}

\begin{figure}[!ht]
    \centering
    \includegraphics[scale=0.55]{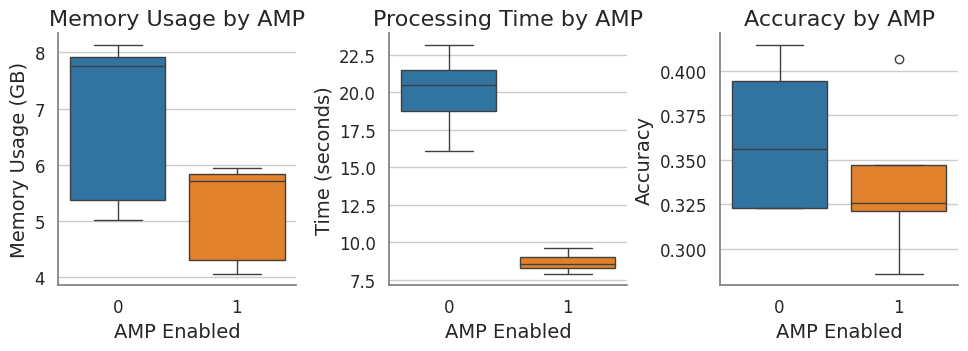}
    \caption{Comparison of Memory Usage, Processing Time, and Accuracy With and Without AMP}
    \label{fig:amp_mem_time_acc}
\end{figure}

\begin{figure}[!ht]
    \centering
    \includegraphics[scale=0.2778]{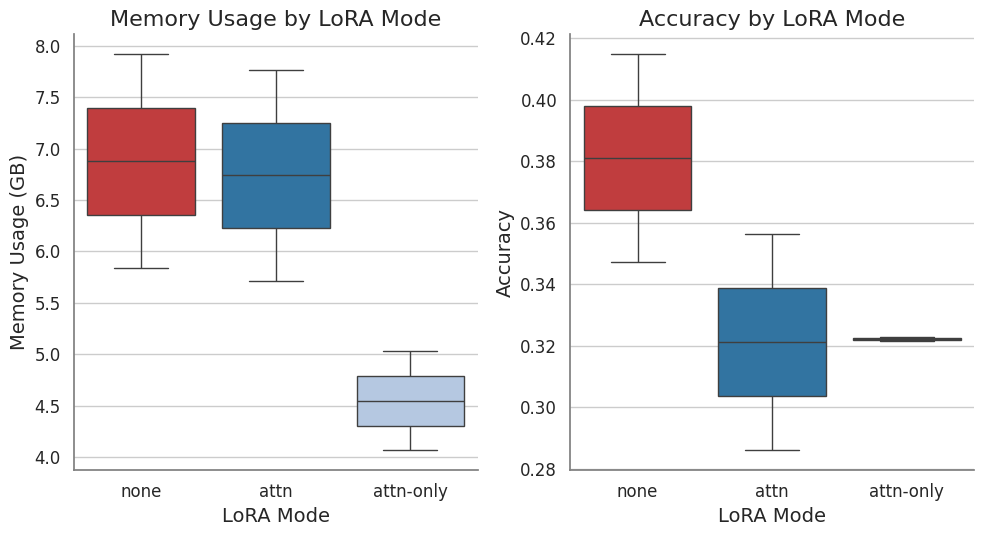}
    \includegraphics[scale=0.2778]{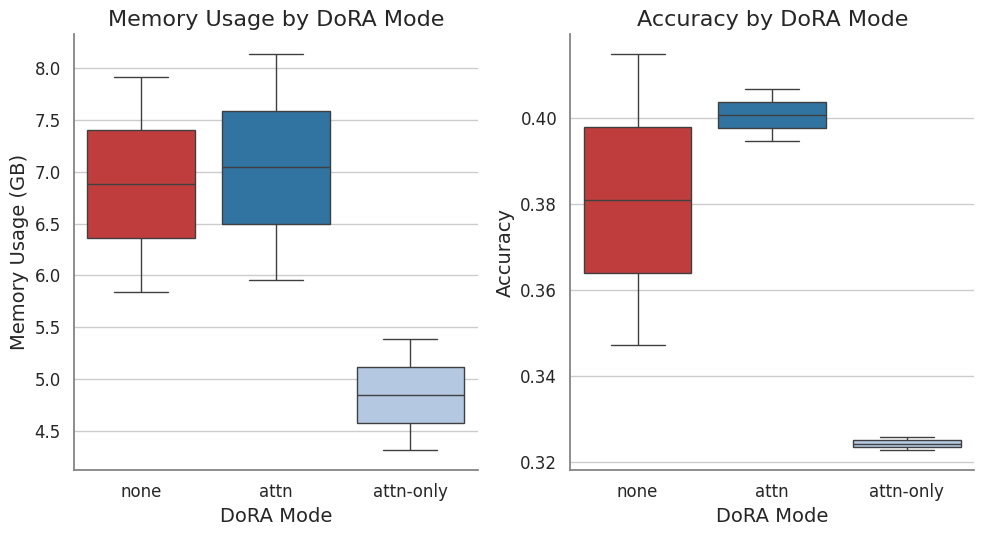}
    \caption{Comparison of Memory Usage and Accuracy Across LoRA/DoRA Modes}
    \label{fig:lora_dora_mem_acc}
\end{figure}
\label{subsec:ablation_details}
\clearpage
\subsection{Final Model Hyperparameters}

\begin{table}[ht]
\centering
\caption{Hyperparameters used for benchmarking the final model configurations.}
\label{tab:final_model_benchmark_hyperparams}
\begin{tabular}{ll}
\toprule
Hyperparameter & Value \\
\midrule
Fine-tune mode & Iterative \\
Learning rate (lr) & $1 \times 10^{-4}$ \\
Batch size & 64 \\
Tasks & SST, Quora, STS \\
Classifier (clf) & Convolutional \\
SST weight decay & $9 \times 10^{-3}$ \\
Paraphrase weight decay & $1 \times 10^{-5}$ \\
STS weight decay & $1 \times 10^{-2}$ \\
Learning rate lambda (lr\_lambda) & 0.5 \\
Optimizer (optim) & Adamax \\
SST learning rate multiplier & 4 \\
Paraphrase learning rate multiplier & 5 \\
STS learning rate multiplier & 3 \\
Epochs & 7 \\
\bottomrule
\end{tabular}
\end{table}

\clearpage

\section{Data Preprocessing Experiments}
\label{sec:data_preprocessing}

In addition to the default string processing included in the handout, we experimented with a modified \texttt{datasets.py} using the \cite{nltk} library to remove stopwords, lemmatize, remove all punctuation, and replaced contractions such as "n't" with "not" and "'ll" with "will". Stopwords are words with low discrimination power, such as ‘a’, ‘an’, ‘the’, ‘and’, etc and their removal has been shown to increase performance on text classification tasks \cite{LadaniDesai2020}. Lemmatization involves breaking a word down into its root meaning (for instance, "best" to "good") and has been shown to improve performance in sentiment analysis tasks \cite{khyani2021}. We hypothesized also that contraction removal could help with text standardization and potentially improve subtask performance. Overall, we did not observe significant differences between our various preprocessing pipelines when applied to model training with our final model architecture (\autoref{sec:best_arch}). Thus, we stuck with only basic white-space removal and special character removal for the models included in our final ensembles (\autoref{sec:final_ensembles}).

\section{Model Parameter Count}
\label{sec:param_ct}

We computed each of our minBERT backbones to contain $\approx 110M$ trainable parameters, and our full set of convolution classifiers to contain $\approx 6M$ trainable parameters. Thus, for the architecture described in \autoref{sec:best_arch}, our model consists of $\approx 226M$ parameters.

\end{document}